\documentclass{ecai}  % use option [doubleblind] for double blind submission and hiding the authors section

\usepackage{graphicx}
\usepackage{latexsym}

\usepackage{amsmath}
\usepackage{amssymb}
\usepackage{mathtools}
\usepackage{graphicx} % DO NOT CHANGE THIS
\usepackage{caption} % DO NOT CHANGE THIS AND DO NOT ADD ANY OPTIONS TO IT
\usepackage{subcaption}
\usepackage{algorithm}
\usepackage{algorithmic}
\usepackage{float}  % we can force table position
\usepackage{booktabs} % for professional tables
\usepackage{longtable} % for big tables.
\usepackage{hyperref}
\newtheorem{thm}{Theorem}
\newtheorem{definition}[thm]{Definition}
\newtheorem{remark}[thm]{Remark}

\usepackage[textsize=tiny]{todonotes}

\newcommand{\R}{\mathbb{R}}

\newcommand{\sta}{\mathcal{S}}
\newcommand{\act}{\mathcal{A}}

\begin{document}

\begin{frontmatter}

\title{Worrisome Properties of Neural Network Controllers\\and Their Symbolic Representations}

%\author[A]{\fnms{Anonymous}~\snm{Authors}}
\author[A]{\fnms{Jacek}~\snm{Cyranka} \thanks{Corresponding Author. Email: jcyranka@gmail.com} }
\author[B]{\fnms{Kevin E M}~\snm{Church}}%\orcid{....-....-....-....}}
\author[C]{\fnms{Jean-Philippe}~\snm{Lessard}}%\orcid{....-....-....-....}} % use of \orcid{} is optional

\address[A]{Institute of Informatics, University of Warsaw}
\address[B]{Centre de Recherches Math\'ematiques, Université de Montréal}
\address[C]{Department of Mathematics and Statistics, McGill University}

\begin{abstract}
We raise concerns about controllers' robustness in simple reinforcement learning benchmark problems. We focus on neural network controllers and their low neuron and symbolic abstractions. A typical controller reaching high mean return values still generates an abundance of persistent low-return solutions, which is a highly undesirable property, easily exploitable by an adversary. We find that the simpler controllers admit more persistent bad solutions. 
We provide an algorithm for a systematic robustness study and prove existence of persistent solutions and, in some cases, periodic orbits, using a computer-assisted proof methodology.
\end{abstract}

\end{frontmatter}

\section{Introduction}\label{Sec:intro}

The study of neural network (NN) robustness properties has a long history in the research on artificial intelligence (AI). Since establishing the existence of so-called adversarial examples in deep NNs in \cite{adversarialexamples}, it is well known that NN can output unexpected results by slightly perturbing the inputs and hence can be exploited by an adversary. Since then, the robustness of other NN architectures has been studied \cite{noderobust}. In the context of control design using reinforcement learning (RL), the robustness of NN controllers has been studied from the adversarial viewpoint \cite{robustadversarial,evaluatingrobustness}. Due to limited interpretability and transparency, deep NN controllers are not suitable for deployment for critical applications. Practitioners prefer abstractions of deep NN controllers that are simpler and human-interpretable. Several classes of deep NN abstractions exist, including single layer or linear nets, programs, tree-like structures, and symbolic formulas. It is hoped that such abstractions maintain or improve a few key features: generalizability -- the ability of the controller to achieve high performance in similar setups (e.g., slightly modified native simulator used in training); deployability -- deployment of the controller in the physical world on a machine, e.g., an exact dynamical model is not specified and the time horizon becomes undefined; verifiability -- one can verify a purported controller behavior (e.g., asymptotic stability) in a strict sense; performance -- the controller reaches a very close level of average return as a deep NN controller. 

In this work, we study the robustness properties of some symbolic controllers derived in \cite{symbolic} as well as deep NN with their a few neuron and symbolic abstractions derived using our methods. By robustness, we mean that a controller maintains its average return values when changing the simulator configuration (scheme/ time-step) at test time while being trained on some specific configuration. Moreover, a robust controller does not admit open sets of simulator solutions with extremely poor return relative to the average. In this regard, we found that NNs are more robust than simple symbolic abstractions, still achieving comparable average return values. To confirm our findings, we implement a workflow of a symbolic controller derivation: regression of a trained deep NN and further fine-tuning. For the simplest benchmark problems, we find that despite the controllers reaching the performance of deep NNs measured in terms of mean return, there exist singular solutions that behave unexpectedly and are persistent for a long time. In some cases, the singular solutions are persistent forever (periodic orbits). The found solutions are stable and an adversary having access to the simulation setup knowing the existence of persistent solutions and POs for specific setups and initial conditions may reconfigure the controlled system and bias it towards the bad persistent solutions; resulting in a significant performance drop, and if the controller is deployed in practice, may even lead to damage of robot/machine. This concern is critical in the context of symbolic controllers, which are simple abstractions more likely to be deployed on hardware than deep NNs. Two systems support the observed issues. First, the standard pendulum benchmark from OpenAI gym \cite{gym} and the cartpole swing-up problem.

Each instance of an persistent solution we identify is verified mathematically using computer-assisted proof (CAP) techniques based on interval arithmetic \cite{MR0231516,Tucker2011}
implemented in Julia \cite{julia}. Doing so, we verify that the solution truly exists and is not some spurious object resulting from e.g., finite arithmetic precision. Moreover, we prove the adversarial exploitability of a wide class of controllers. The existence of persistent solutions is most visible in the case of symbolic controllers. For deep NN, persistent solutions are less prevalent, and we checked that deep NN controllers' small NN abstractions (involving few neurons) somewhat alleviate the issue of symbolic controllers, strongly suggesting that the robustness is inversely proportional to the number of parameters, starkly contrasting with common beliefs and examples in other domains. 
%
%
%

%\paragraph{Main Contributions.}
\vspace{.2cm} {\bf Main Contributions.}
Let us summarize the main novel contributions of our work to AI community below.

\emph{Systematic controller robustness study.} In light of the average return metric being sometimes deceptive, we introduce a method for investigating controller robustness by designing an persistent solutions search and the penalty metric.

\emph{Identification and proofs of abundant persistent solutions.} We systematically find and prove existence of a concerning number of persistent orbits for symbolic controllers in simple benchmark problems. Moreover, we carried out a proof of a periodic orbit for a deep NN controller, which is of independent interest. To our knowledge, this is the first instance of such a proof in the literature.

\emph{NN controllers are more robust than symbolic.} We find that the symbolic controllers admit significantly more bad persistent solutions than the deep NN and small distilled NN controllers.
\subsection{Related Work}
\textit{(Continuous) RL.} A review of RL literature is beyond the scope of this paper (see \cite{sutton} for an overview). In this work we use state-of-the-art TD3 algorithms dedicated for continuous state/action spaces \cite{td3} based on DDPG \cite{ddpg}. Another related algorithm is SAC \cite{sacc}. 

\textit{Symbolic Controllers.} Symbolic regression as a way of obtaining explainable controllers appeared in \cite{kubalik,hein,symbolic}. Other representations include programs \cite{verma,synthesizeprograms} or decision trees \cite{towardinterpretable}. For a broad review of explainable RL see \cite{xaireview}.

\textit{Falsification of Cyber Physical Systems (CPS)} The research on falsification \cite{fals3,fals4,fals1,fals2} utilizes similar techniques for demonstrating the violation of a temporal logic formula, e.g., for finding solutions that never approach the desired equilibrium. We are interested in solutions that do not reach the equilibrium but also, in particular, the solutions that reach minimal returns. 

\textit{Verification of NN robustness using SMT} Work on SMT like ReLUplex \cite{smt2,smt1,smt3} is used to construct interval robustness bounds for NNs only. In our approach we construct interval bounds for solutions of a coupled controller (a NN) with a dynamical system and also provide existence proofs.

\textit{Controllers Robustness.} Design of NN robust controllers focused on adversarial defence methods \cite{robustadversarial,evaluatingrobustness}.

\textit{CAPs.} Computer-assisted proofs for ordinary differential equations (ODEs) in AI are not common yet. Examples include validation of NN dynamics \cite{computervalidation} and proofs of spurious local minima \cite{spuriouslocal}.
\subsection{Structure of the Paper}
Section \ref{sec-prelim} provides background on numerical schemes and RL framework used in this paper. Section \ref{sec-alg} describes the training workflow for the neural network and symbolic controllers. The class of problems we consider is presented in Section \ref{sec-problems}. We describe the computer-assisted proof methodology in Section \ref{sec:cap_methodology}. Results on persistent periodic orbits appear in Section \ref{sec:motivating_example}, and we describe the process by which we search for these and related singular solutions in Section \ref{sec::robustness}.
\section{Preliminaries}\label{sec-prelim}
\subsection{Continuous Dynamics Simulators for AI}
% In this section, we present the construction of simulators used for controller training procedures via AI techniques, which is the basis of our robustness study. 
Usually, there is an underlying continuous dynamical system with control input that models the studied problem $s^\prime(t) = f(s(t), a(t))$, where $s(t)$ is the state, $a(t)$ is the control input at time $t$, and $f$ is a vector field. For instance, the rigid body general equations of motion in continuous time implemented in robotic simulators like MuJoCo \cite{todorov2012mujoco} are $Mv^\prime+c = \tau + J^Tf$, $J,f$ is the constraint Jacobian and force, $\tau$ is the applied force, $M$ inertia matrix and $c$ the bias forces. For training RL algorithms, episodes of simulated rollouts $(s_0,a_0,r_1,s_1,\dots)$ are generated; the continuous dynamical system needs to be discretized using one of the available numerical schemes like the Euler or Runge-Kutta schemes \cite{solvingodes}. After generating a state rollout, rewards are computed $r_{k+1}=r(s_k, a_k)$. The numerical schemes are characterized by the approximation order, time-step, and explicit/implicit update. In this work, we consider the explicit Euler (E) scheme $s_{k+1} = s_k + hf(s_k, a_k)$; this is a first-order scheme with the quality of approximation being proportional to time-step $h$ (a hyperparameter). Another related scheme is the so-called semi-implicit Euler (SI) scheme, a two-step scheme in which the velocities are updated first. Then the positions are updated using the computed velocities. Refer to the appendix for the exact form of the schemes.

In the research on AI for control, the numerical scheme and time-resolution\footnote{While in general time-resolution may not be equal to the time step, in this work we set them to be equal.
%; e.g., a variable time step or several smaller steps may be used to generate the following observation. However, this work will consider cases exclusively when they are equal. 
} of observations $h$ are usually fixed while simulating episodes. 
Assume we are given a controller that was trained on simulated data generated by a particular scheme and $h$; we are interested in studying the controller robustness and properties after the zero-shot transfer to a simulator utilizing a different scheme or $h$, e.g., explicit to semi-implicit or using smaller $h$'s. %In the sequel we define a family of Markov Decision Processes (MDPs) with different transition functions. 
\subsection{Reinforcement Learning Framework}
Following the standard setting used in RL, we work with a Markov decision process (MDP) formalism $(\sta,\act, F, r, \rho_0, \gamma)$, where $\sta$ is a state space, $\act$ is an action space, $F\colon \sta\times\act\to\sta$ is a deterministic transition function, $r\colon\sta\times\act\to\R$ is a reward function, $\rho_0$ is an initial state distribution, and $\gamma\in(0,1)$ is a discount factor used in training. $\sta$ may be equipped with an equivalence relation, e.g. for an angle variable $\theta$, we have $\theta\equiv\theta + k2\pi$ for all $k\in\mathbb{Z}$. In RL, the agent (policy) interacts with the environment in discrete steps by selecting an action $a_t$ for the state $s_t$ at time $t$, causing the state transition $s_{t+1}=F(s_t,a_t)$; as a result, the agent collects a scalar reward $r_{t+1}(s_t,a_t)$, the (undiscounted) return is defined as the sum of discounted future reward $R_t = \sum_{i=t}^T{r(s_i,a_i)}$ with $T>0$ being the fixed episode length of the environment.  
RL aims to learn a policy that maximizes the expected return over the starting state distribution. 

In this work, we consider the family of MDPs in which the transition function is a particular numerical scheme. We study robustness w.r.t. the scheme; to distinguish the \emph{transition function used for training (also called native)} from the \emph{transition function used for testing}, we introduce the notation $F_{train}$ and $F_{test}$ resp. e.g. explicit Euler with time-step $h$ is denoted $F_*({\rm E},h)$, where $* \in \{test,train\}$.

\section{Algorithm for Training of Symbolic Controllers and Small NNs}\label{sec-alg}
Carrying out the robustness study of symbolic and small NN controllers requires that the controllers are first constructed (trained). We designed a three-step deep learning algorithm for constructing symbolic and small NN controllers. Inspired by the preceding work in this area the controllers are derived from a deep RL NN controller. The overall algorithm is summarized in Alg.~\ref{alg:symbolic}.
%
%\vspace{-.2cm}
%
\begin{algorithm}[h!]
{\small
   \caption{Symbolic/Small NN Controllers Construction}
   \label{alg:symbolic}
\begin{algorithmic}[1]
   \INPUT MDP determining studied problem; RL training $h$-params; symbolic \& small NN regression $h$-params; fine-tuner $h$-params;
   \OUTPUT deep NN policy $\pi_{deep}$; small NN policy $\pi_{small}$; family of symbolic policies $\{\pi_{symb, k}\}$ ($k$ complexity);
   \STATE Apply an off-policy RL algorithm for constructing a deterministic deep NN policy $\pi_{deep}$;
   \STATE Using the replay buffer data apply symbolic regression for computing symbolic abstractions $\{\pi_{symb, k}\}$ (having complexity $k$) of deep NN controller and MSE regression for small NN $\pi_{small}$ policy distillation;
   \STATE Fine-tune the constructed controllers parameters for maximizing the average return using CMA-ES and/or analytic gradient.
\end{algorithmic}}
\end{algorithm}
%
%\vspace{-.5cm}
%
\subsection{RL Training}
First we train a deep NN controller using the state-of-the-art model-free RL algorithm TD3 \cite{ddpg,td3} -- the SB3 implementation \cite{sb3}. We choose TD3, as it utilizes the replay buffer and constructs deterministic policies (NN). Plots with the evaluation along the training procedure for studied systems can be found in App.~\ref{sec:training}. 
\subsection{Symbolic Regression}
A random sample of states is selected from the TD3 training replay buffer. Symbolic abstractions of the deep NN deterministic policies are constructed using the symbolic regression over the replay buffer samples. Following earlier work \cite{kubalik,hein,symbolic} the search is performed by an evolutionary algorithm. For such purpose, we employ the PySR Python library \cite{pysr,cranmer2020discovering}. 
% It is flexible and fast, as its backbone is written in Julia. 
The main hyperparameter of this step is the complexity limit (number of unary/binary operators) of the formulas ($k$ in Alg.~\ref{alg:symbolic}). 
This procedure outputs a collection of symbolic representations with varying complexity. Another important hyperparameter is the list of operators used to define the basis for the formulas. We use only the basic algebraic operators (add, mul., div, and multip.\ by scalar). We also tried a search involving nonlinear functions like $tanh$, but the returns were comparable with a larger complexity. 
\subsection{Distilling Simple Neural Nets}
Using a random sample of states from the TD3 training replay buffer we find the parameters of the small NN representation using the mean-squared error (MSE) regression. 
\subsection{Controller Parameter Fine-tuning}
Just regression over the replay buffer is insufficient to construct controllers that achieve expected returns comparable with deep NN controllers, as noted in previous works. The regressed symbolic controllers should be subject to further parameter fine-tuning to maximize the rewards. There exist various strategies for fine-tuning. In this work, we use the non-gradient stochastic optimization covariance matrix adaptation evolution strategy (CMA-ES) algorithm \cite{cmaes,pycma}. 
% We find CMA-ES the most suitable for the fine-tuning task. 
We also implemented analytic gradient optimization, which takes advantage of the simple environment implementation, and performs parameter optimization directly using gradient descent on the model rollouts from the differentiable environment time-stepping implementation in PyTorch. 
% However, we find it to work only for the pendulum.
%
%
%
\section{Studied Problems}\label{sec-problems}
We perform our experimental investigation and CAP support in the setting of two control problems belonging to the set of standard benchmarks for continuous optimization. 
% including RL and evolutionary search. 
First, the pendulum problem is part of the most commonly used benchmark suite for RL -- OpenAI gym \cite{gym}. Second, the cart pole swing-up problem is part of the DeepMind control suite \cite{dmcontrol}. Following the earlier work \cite{pilco} we used a closed-form implementation of the cart pole swing-up problem. While these problems are of relatively modest dimension, compared to problems in the MuJoCo suite, we find them most suitable to convey our message. 
% These are widely accepted benchmarks by the AI community. 
The low system dimension makes a self-contained cross-platform implementation easier and eventually provides certificates for our claims using interval arithmetic and CAPs. 
\subsection{Pendulum}
The pendulum dynamics is described by a 1d $2^{nd}$ order nonlinear ODE. We followed the implementation in OpenAI gym, where the ODEs are discretized with a semi-implicit (SI) Euler method with $h=0.05$. For training we use $F_{train}({\rm SI}, 0.05)$. Velocity $\omega$ is clipped to the range $[-8,8]$, and control input $a$ to $[-2,2]$. There are several constants: gravity, pendulum length and mass $(g,l,m)$, which we set to defaults. See App.~\ref{sec:pendulum} for the details. The goal of the control is therefore to stabilize the up position $\theta = 0\mod 2\pi$, with zero angular velocity $\omega$. The problem uses quadratic reward for training and evaluation $r = -\lfloor\theta\rfloor^2 - 0.1\omega^2-0.001a^2$, where  $\lfloor\theta\rfloor=\arccos(\cos(\theta))$ at given time $t$ and action $a$. The episode length is $200$ steps. The max reward is $0$, and large negative rewards might indicate long-term simulated dynamics that are not controlled. 
% For example, the presence of stable periodic orbits is consistent with linear scaling of the mean reward with respect to the episode length. 
% {\color{blue}If the controller were to successfully stabilize $\theta=0$, then we would expect sub-linear scaling of the reward, given that the dynamics close to the equilibrium should be linear-dominated and therefore exponentially damped. This latter remark, however, assumes that the controller applies zero torque at $(\omega,\theta)=(0,0)$.}
%
%
%
\subsection{Cartpole Swing-up}
The cartpole dynamics is described by a 2d $2^{nd}$ order nonlinear ODEs with two variables: movement of the cart along a line ($x,x^\prime)$, and a pole attached to the cart $(\theta, \theta^\prime)$.   We followed the implementation given in \cite{hacartpole}. The ODEs are discretized by the explicit Euler (E) scheme with $h=0.01$. As with the pendulum we use clipping on some system states, and several constants are involved, which we set to defaults. See~\ref{sec:cartpole} for details. The goal of the control is to stabilize the pole upwards $\theta = 0 \mod 2\pi$ while keeping the cart $x$ within fixed boundaries. The problem uses a simple formula for reward $r=\cos{\theta}$, plus the episode termination condition if $|x|$ is above threshold. The episode length is set to $500$, hence the reward is within $[-500,500]$. 
% , but unlikely value close to the either limit. {\color{blue} For detecting transients, we use a modified reward function, disabling the termination condition, and subtracting an additional penalty for the velocity. 
Large negative reward is usually indicative of undesirable behaviour, with the pole continuously oscillating, the cart constantly moving, and escaping the boundaries fairly quickly.
\section{Rigorous Proof Methodology}\label{sec:cap_methodology}
All of our theorems presented in the sequel are supported by a computer-assisted proof, guaranteeing that they are fully rigorous in a mathematical sense. Based on the existing body of results and our algorithm we developed in Julia, we can carry out the proofs for different abstractions and problems as long as the set of points of non-differentiability is small, e.g., it works for almost all practical applications: ReLU nets, decision trees, and all sorts of problems involving dynamical systems in a closed form. The input to our persistent solutions prover is a function in Julia defining the controlled problem, the only requirement being that the function can be automatically differentiated. To constitute a proof, this part needs to be carried out rigorously with interval arithmetic. Our CAPs are automatic; once our searcher finds a candidate for a persistent solution/PO, a CAP program attempts to verify the existence of the solution/PO by verifying the theorem (Theorem~\ref{theorem-radpol}) assumptions. If the prover succeeds this concludes the proof.
\subsection{Interval Arithmetic}
\begin{figure}[h!]
\centering
\begin{subfigure}[t]{0.45\columnwidth}
\includegraphics[scale=0.21]{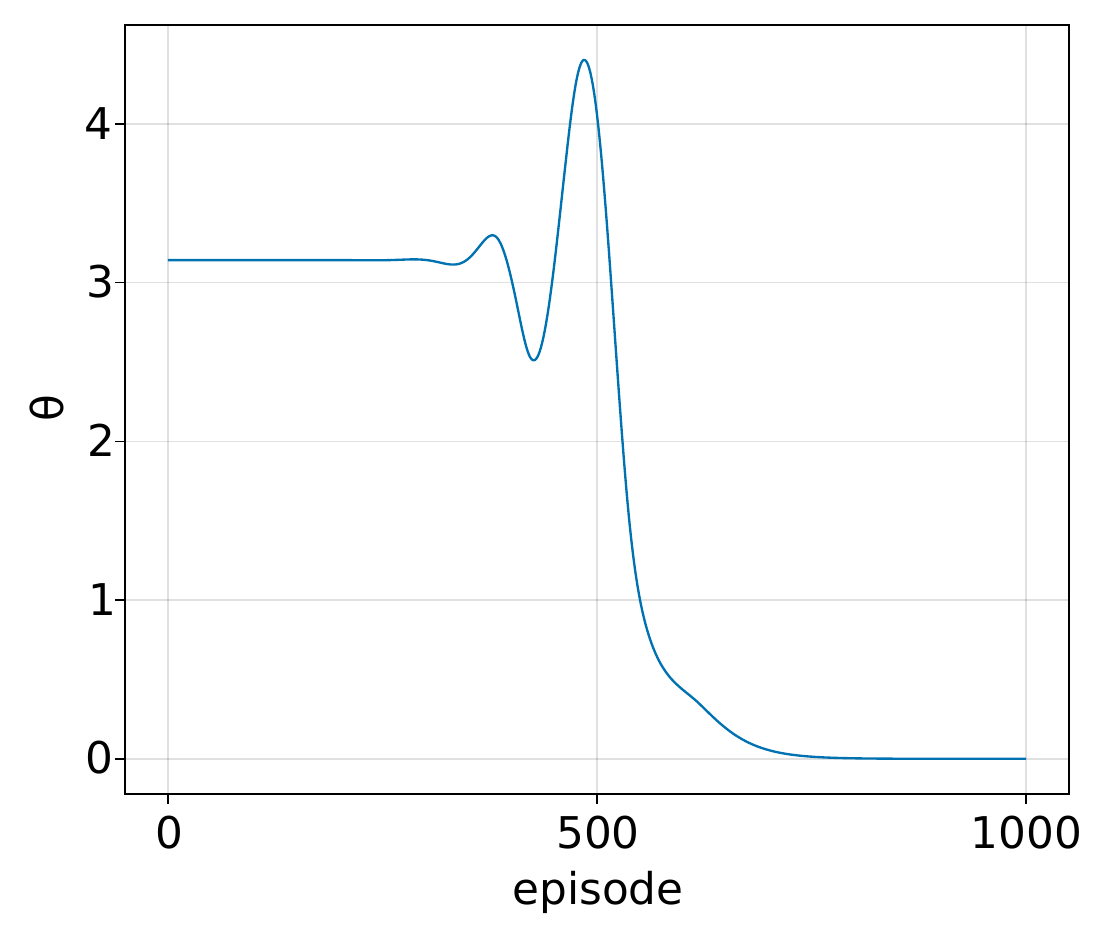}
\end{subfigure}
\begin{subfigure}[t]{0.45\columnwidth}
\includegraphics[scale=0.21]{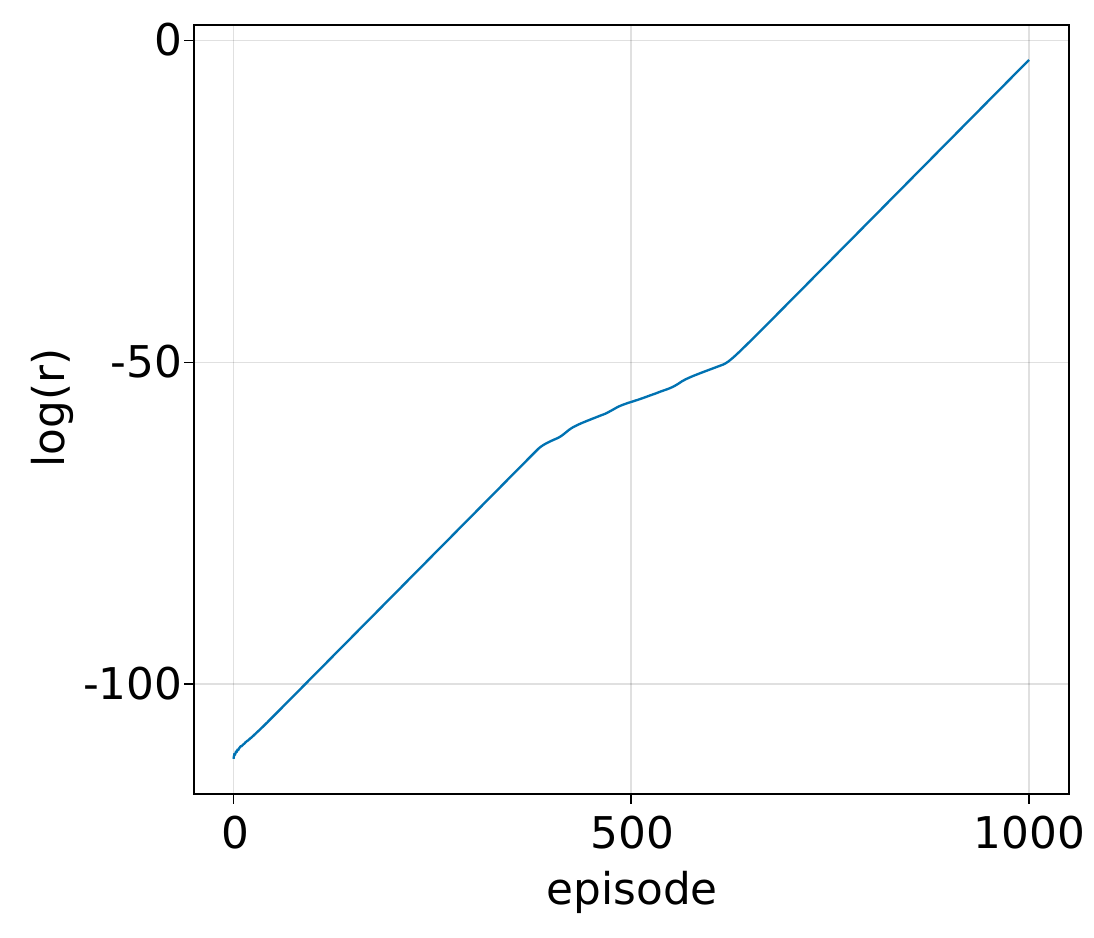}
\end{subfigure}
%\vspace{-.2cm}
\caption{Left: midpoint of interval enclosure of a proven persistent solution (see Appendix Tab.\ \ref{9A_CMA_transients}). Right: log-scale of radius of the interval enclosure. Calculations done at 163 bit precision, the minimum possible for this solution at episode length 1000.}\label{fig:transient2_wrapping}
\end{figure}

Interval arithmetic is a method of tracking rounding error in numerical computation. Operations on floating point numbers are instead done on \textit{intervals} whose boundaries are floating point numbers. 
% For example, multiplication of intervals $[a,b]$ and $[c,d]$ is defined to be the smallest interval containing the set of products, $\{xy : x\in[a,b],y\in[c,d]\}$, and whose boundaries are floating point numbers. 
Functions $f$ of real numbers are \textit{extended} to functions $\overline{f}$ defined on intervals, with the property that $\overline{f}(X)$ necessarily contains $\{f(x):x\in X\}.$ The result is that if $y$ is a real number and $Y$ is a thin interval containing $y$, then $f(y)\in \overline f(Y)$. For background, the reader may consult the books \cite{MR0231516,Tucker2011}. Function iteration on intervals leads to the \textit{wrapping effect}, where the radius of an interval increases along with composition depth. See Figure \ref{fig:transient2_wrapping} for a visual.

\subsection{Computer-assisted Proofs of Periodic Orbits}\label{sec:CAP}
% The main idea behind our computer-assisted proofs of periodic orbits is a particular twist on Newton's Method. 
For $x=(x_1,\dots,x_n)$, let $||x||=\max\{|x_1|,\dots,|x_n|\}$. The following is the core of our CAPs.
\begin{theorem}\label{theorem-radpol}
Let $G:U\rightarrow \mathbb{R}^n$ be continuously differentiable, for $U$ an open subset of $\mathbb{R}^n$. Let $\overline x\in\mathbb{R}^n$ and $r^*\geq 0$. Let $A$ be a $n\times n$ matrix\footnote{In practice, a numerical approximation $A\approx DF(\overline x)^{-1}$.} of full rank. Suppose there exist real numbers $Y$, $Z_0$ and $Z_2$ such that
%
%\vspace{-.1cm}
\begin{align}%\allowdisplaybreaks
    \label{Y-bound}||AG(\overline x)||&\leq Y,\\
    \label{Z0-bound}||I-ADG(\overline x)||&\leq Z_0\\
    \label{Z2-bound}\sup_{|\delta|\leq r^*}||A(DG(\overline x + \delta) - DG(\overline x))||&\leq Z_2,
\end{align}
%
%\vspace{-.1cm}
where $DG(x)$ denotes the Jacobian of $G$ at $x$, and the norm on matrices is the induced matrix norm. If $Z_0+Z_2<1$ and $Y/(1-Z_0-Z_2)\leq r_*$, the map $G$ has a unique zero $x$ satisfying $||x-\overline x||\leq r$ for any $r\in(Y/(1-Z_0-Z_2),r_*]$.
\end{theorem}
A proof can be completed by following Thm~2.1 in \cite{Day2007}. In Sec.~\ref{sec:map}, we identify $G$ whose zeroes correspond to POs. 
% This is how Theorem \ref{theorem-radpol} is used to prove the existence of periodic orbits. 
Conditions \eqref{Y-bound}--\eqref{Z2-bound} imply that the Newton-like operator
$T(x) = x - AG(x)$
is a contraction on the closed ball centered at the \textit{approximate zero} $\overline x$ with radius $r>0$. Being a contraction, it has a unique fixed point ($x$ such that $x = T(x)$) by the Banach fixed point theorem. As $A$ is full rank,  $G(x)=0$, hence an orbit exists. The radius $r$ measures how close the approximate orbit $\overline x$ is to the exact orbit, $x$. The contraction is rigorously verified by performing all necessary numerical computations using interval arithmetic. The technical details appear in App.~\ref{sec:Julia_implemtation_details}.
% \begin{remark}
% In applications, the matrix $A$ is usually taken to be the inverse of $DF(\overline x)$. In the computer implementation, matrix inversion is almost never exact, so $A$ is a machine-computed approximate inverse. As such, $Z_0$ is a measure of quality of the approximation $DF(\overline x)^{-1}\approx A$.
% \end{remark}
%\todo[inline]{general presentation of bad solutions in considered classical control problems}
%\subsection{Quasi-periodic Orbits}

\subsection{Set-up of the Nonlinear Map}\label{sec:map}
A PO is a finite MDP trajectory. Let the step size be $h$, and let the period of the orbit be $m$. We present a nonlinear map that encodes (as zeroes of the map) POs when $h$ is fixed. However, for technical reasons (see App.~\ref{sec:discretized_qpo}), it is possible for such a proof to fail. If Alg.~\ref{algo_2} fails to prove the existence of an orbit with a fixed step size $h$, we fall back to a formulation where the step size is not fixed, which is more likely to yield a successful proof. This alternative encoding map $G_2$ is presented in App.~\ref{sec:variable-step}.
Given $h$, pick $g(h,\cdot) \in \{ g_{\rm E},g_{\rm SI}\}$ one of the discrete dynamical systems used for numerically integrating the ODE. Let $p$ be the dimension of the state space, so $g(h,\cdot):\mathbb{R}^p\rightarrow\mathbb{R}^p$. We interpret the first dimension of $\mathbb{R}^p$ to be the angular component, so that a periodic orbit requires a shift by a multiple of $2\pi$ in this variable. Given $h$, the number of steps $m$ (i.e.\ period of the orbit) and the number of signed rotations $j$ in the angular variable, POs are zeroes of the map (if and only if) $G_1:\mathbb{R}^{pm}\rightarrow\mathbb{R}^{pm}$, defined by 
\[
%{\footnotesize
G_1(X) = \begin{pmatrix}
x_0-g(h,x_{m})+(j2\pi,\mathbf{0})
\\
x_1-g(h,x_{0})
\\
x_2-g(h,x_{1})
\\
\vdots
\\
x_m-g(h,x_{m-1})
\end{pmatrix},%}
\]
where $\mathbf{0}$ is the zero vector in $\mathbb{R}^{p-1}$, $X=(x_1,\dots,x_m)$ for $x_i\in\mathbb{R}^p$, and $x_1,\dots,x_m$ are the time-ordered states. % of the orbit.
% This formulation is used when we want to prove a simulated orbit at a specific step size (e.g.\ $0.05$, $0.01$, $0.005$ or other common step sizes). 
% Note that $j$ represents the number of times the pendulum rotates counterclockwise around the pivot point.
%
%
%
\section{Persistent Orbits in 
Controlled Pendulum}\label{sec:motivating_example}
When constructing controllers using machine learning or statistical methods, the most often used criterion for measuring their quality is the mean return from performing many test episodes. The mean return may be a deceptive metric for constructing robust controllers. More strongly, our findings suggest that mean return is not correlated to the presence of periodic orbits or robustness.  One would typically expect a policy with high mean return to promote convergence toward states that maximize the return for any initial condition (IC) and also for other numerical schemes. Our experiments revealed reasons to believe this may be true for deep NN controllers. However, in the case of simple symbolic controllers, singular persistent solutions exist that accumulate large negative returns at a fast pace. By persistent solutions we mean periodic orbits that remain $\varepsilon$ away from the desired equilibrium. This notion we formalize in Sec.~\ref{sec:persistent}. We emphasize that all of the periodic orbits that we prove are necessary stable in the usual Lyapunov sense, i.e., the solutions that start out near an equilibrium stay near the equilibrium forever, and hence feasible in numerical simulations.
We find such solutions for controllers as provided in the literature and constructed by ourselves employing Alg.~\ref{alg:symbolic}. We emphasize that our findings are not only numerical, but we support them with (computer-assisted) mathematical proofs of existence. 
%u
%
%
\subsection{Landajuela et.\ al \cite{symbolic} Controller}
\label{sec:landuleja_controller}
\begin{figure}[h!]
     \centering
     \captionsetup[subfigure]{justification=centering}
     \begin{subfigure}[t]{0.325\columnwidth}
         \centering
         \includegraphics[width=\textwidth]{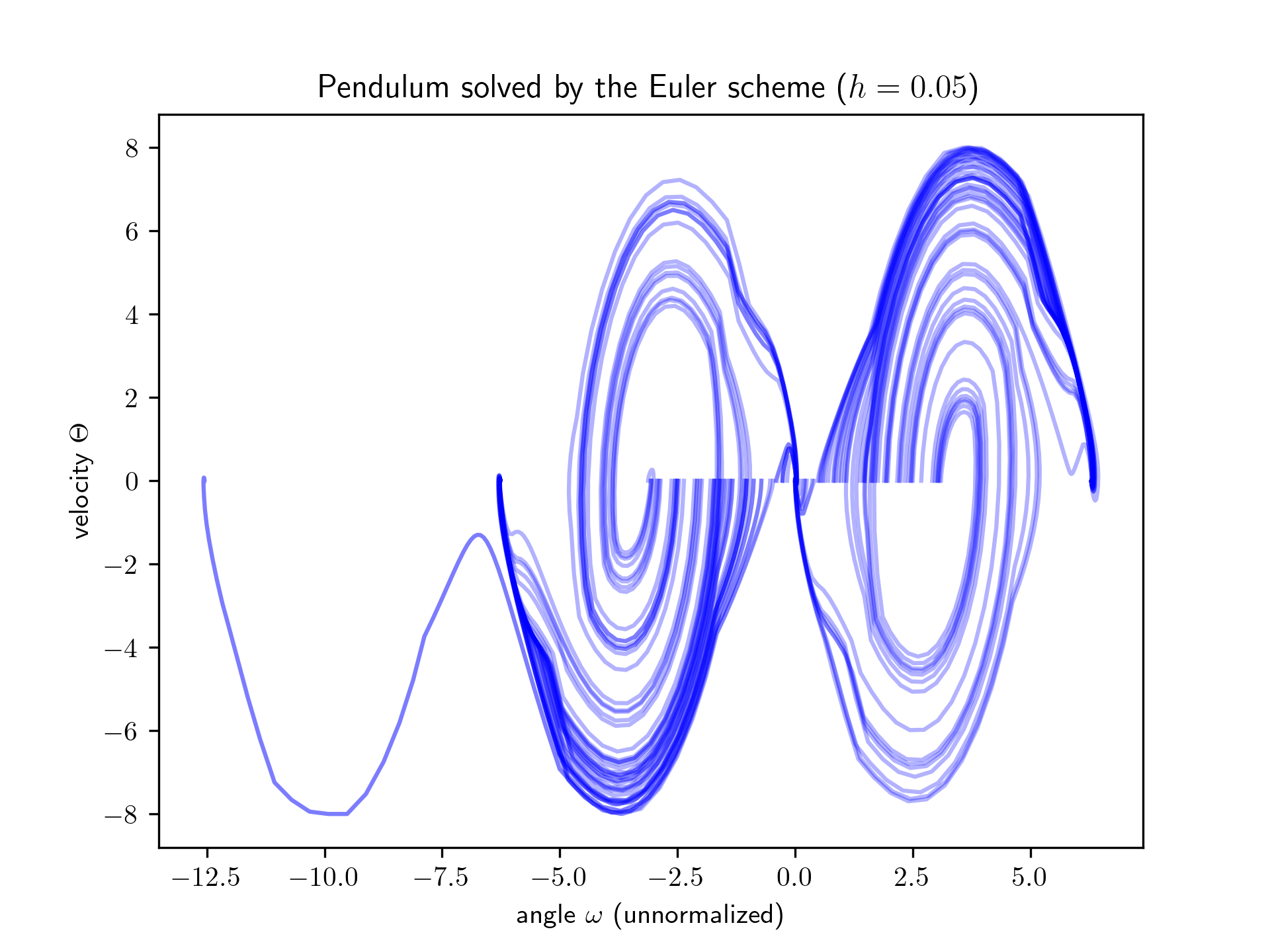}
         \captionsetup{justification=centering}
         \caption{(SI), $h=0.05$ (native)}
         \label{fig:sim1}
     \end{subfigure}
     %\hfill
     \begin{subfigure}[t]{0.325\columnwidth}
         \centering
         \includegraphics[width=\textwidth]{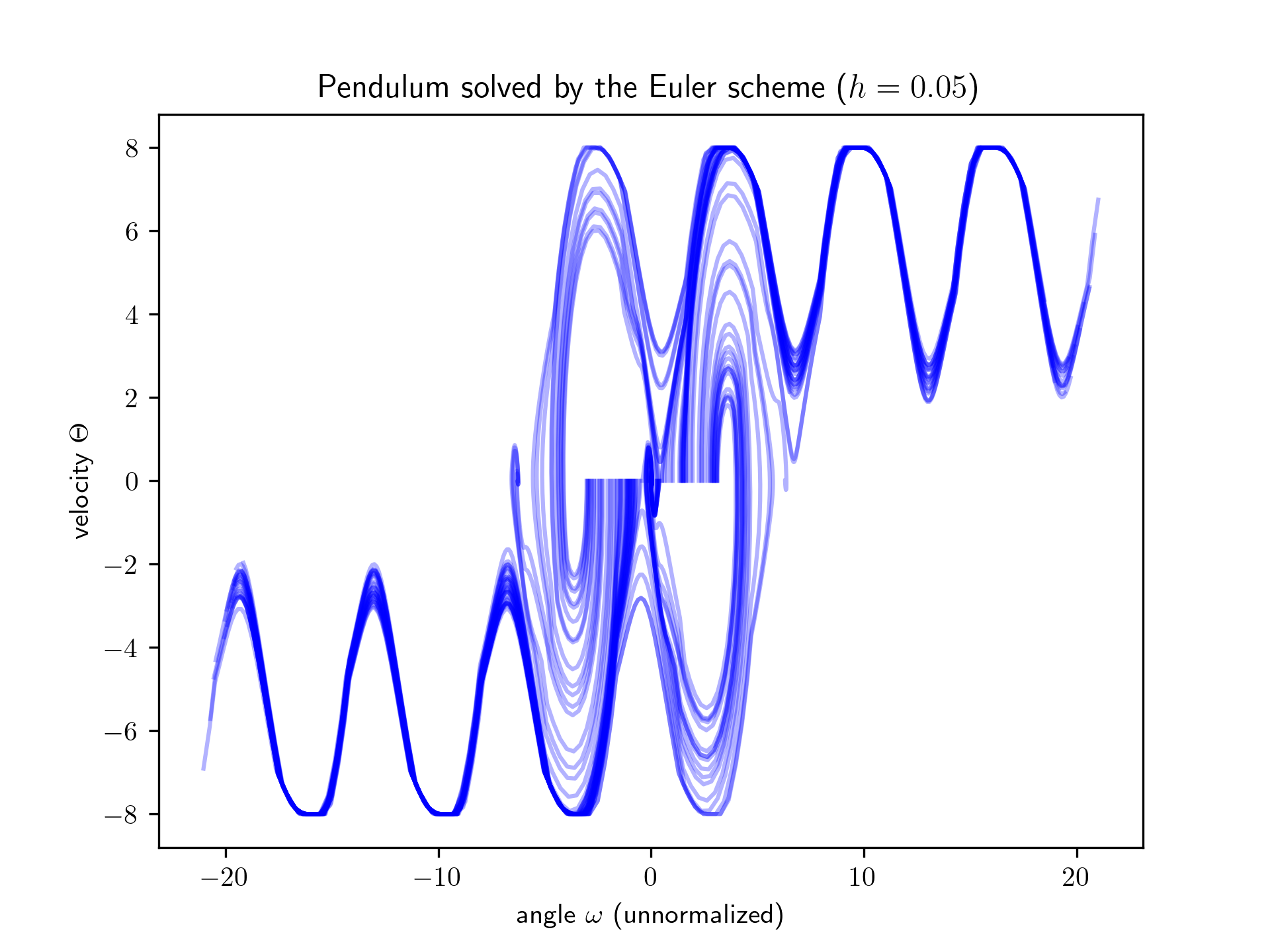}
         \captionsetup{justification=centering}
         \caption{(E), $h=0.05$ }
         \label{fig:sim2}
     \end{subfigure}
     \begin{subfigure}[t]{0.325\columnwidth}
         \centering
         \includegraphics[width=\textwidth]{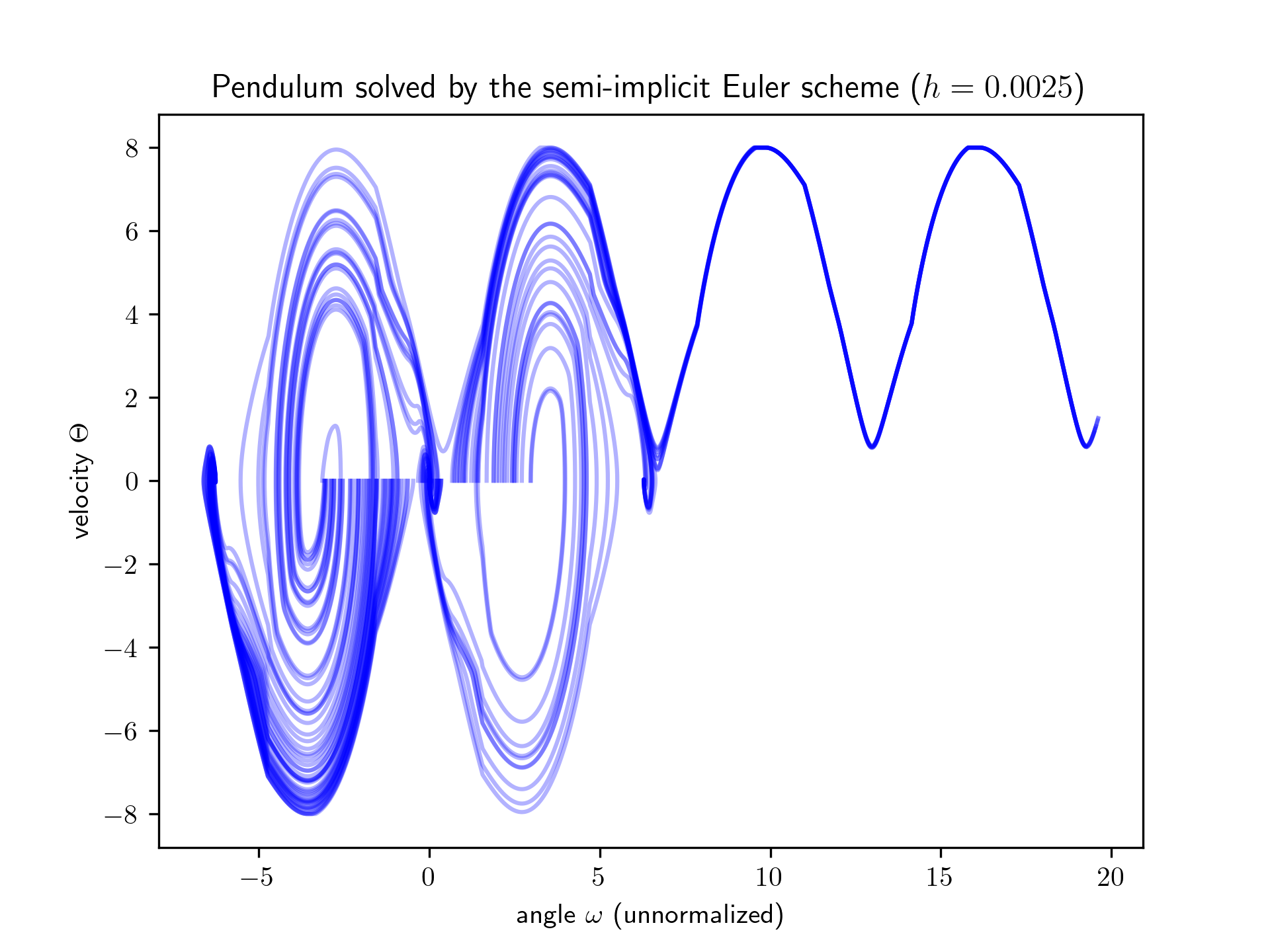}
         \captionsetup{justification=centering}
         \caption{(SI), $h=0.025$}
         \label{fig:sim3}
     \end{subfigure}
     %\vspace{-.2cm} 
        \caption{$100$ numerical simulations with IC $\omega=0$ and $\theta$ sampled uniformly, time horizon set to $T=6$, $x$-axis shows the (unnormalized) $\omega$, and $y$-axis $\theta$. In (a), all IC are attracted by an equilibrium at $\omega=0\mbox{mod} 2\pi$, $\theta=0$. Whereas when applying different $F_{test}$, (b) and (c) show existence of attracting periodic solutions (they can be continued infinitely as our theorems demonstrate).}
        %\vspace{-10pt}
        \label{fig:simulations}
\end{figure}

First, we consider the symbolic low complexity %(yet achieving high returns) 
controller for the pendulum $a = -7.08s_2 - (13.39s_2 + 3.12s_3) / s_1 + 0.27$, derived in \cite{symbolic} (with model given in App.~\ref{sec:pendulum}), where $s_1 = \cos{\theta}$, $s_2 = \sin{\theta}$, $s_3 = \omega = \theta^\prime$, and $a$ is the control input. While this controller looks more desirable than a deep NN with hundreds thousand of parameters, its performance changes dramatically when using slightly different transition function at test-time, i.e., halved $h$ ($F_{test}({\rm SI}, 0.025)$) or the explicit Euler scheme ($F_{test}({\rm E}, 0.05)$). Trajectories in Fig.~\ref{fig:simulations} illustrate that some orbits oscillate instead of stabilizing at the equilibrium $\hat{s}=\hat{\theta}=0 \text{ mod }2\pi$. The average return significantly deteriorates for the modified schemes and the same ICs compared to $F_{train}({\rm SI}, 0.05)$; see Tab.~\ref{tab:symbolic}. Such issues are present in deep NN controllers and small distilled NN to a significantly lower extent.
We associate the cause of the return deterioration with existence of 'bad' solutions -- persistent periodic orbits (POs) (formal Def.~\ref{def:periodic}). Using CAPs (c.f., Sec.~\ref{sec:cap_methodology}) we obtain:

\begin{theorem}\label{thm-land}
For $h \in H=\{0.01, 0.005, 0.0025, 0.001\}$, the nonlinear pendulum system with controller $a$ from \cite{symbolic} described in the opening paragraph of Section \ref{sec:landuleja_controller} has a periodic orbit (PO) under the following numerical schemes;

%\vspace{-.05cm}

1) {\rm (SI)} with step size $h\in H$,

%\vspace{-.17cm}    
2) {\rm (E)} at $h=0.05$ (native), and for all $h\in H$.

%\vspace{-.1cm}  
The identified periodic orbits are persistent (see Def.~\ref{def:persistent}) and generate minus infinity return for infinite episode length, with each episode decreasing the reward by at least 0.198.
\end{theorem}
%
%\vspace{-.2cm} 

%

\subsection{Our Controllers}
The issues with robustness and performance of controllers of Sec.~\ref{sec:landuleja_controller} may be an artefact of a particular controller construction rather than a general property. Indeed, that controller had a division by $s_1$. To investigate this further we apply Alg.~\ref{alg:symbolic} for constructing symbolic controllers of various complexities (without divisions). Using Alg.~\ref{alg:symbolic} we distill a small NN (single hidden layer with $10$ neurons) for comparison. In step 2 we use fine-tuning based on either analytic gradient or CMA-ES, each leading to different controllers. The studied controllers were trained using the default transition $F_{train}({\rm SI},0.05)$, and for testing using $F_{test}({\rm E}, 0.05)$, $F_{test}({\rm E}, 0.025)$, $F_{test}({\rm SI}, 0.05)$, $F_{test}({\rm SI}, 0.025)$.

Tab.~\ref{tab:symbolic} reveals that the average returns deteriorate when using other numerical schemes for the symbolic controllers obtained using Alg.~\ref{alg:symbolic}, analogous to the controller from \cite{symbolic}. The average return discrepancies are very large as well. We emphasize that all of the studied metrics for the symbolic controllers are far from the metrics achieved for the deep NN controller. Terminating Alg.~\ref{alg:symbolic} at step 2 results in a very bad controller achieving mean return only of $-1061$, i.e., as observed in the previous works the symbolic regression over a dataset sampled from a trained NN is not enough to construct a good controller. Analogous to Theorem \ref{thm-land}, we are able to prove the following theorems on persistent periodic orbits (Def.~\ref{def:periodic}) for the controllers displayed in Table \ref{tab:symbolic}.

\begin{theorem}\label{thm-9A-AG}
For $h \in H=\{0.025, 0.0125\}$, the nonlinear pendulum system with controller generated by analytic gradient refinement in Tab.~\ref{tab:symbolic} has POs under

%\vspace{-.07cm}

1) ({\rm SI}) with $h\in H$ and at the native step size $h=0.05$,
%\vspace{-.19cm}   

2) ({\em E}) with $h\in H$.

%\vspace{-.12cm}  
    
The identified periodic orbits are persistent (see Def.~\ref{def:persistent}) and generate minus infinity return for infinite episode length, with each episode decreasing the reward by at least $0.18$.
\end{theorem}

\begin{theorem}\label{thm-9A-CMA-ES}
For $h=0.0125$ and $h=0.05$ (native) with scheme ({\rm E}), the nonlinear pendulum system with controller generated by CMA-ES refinement in Tab.~\ref{tab:symbolic} has POs which generate minus infinity return for infinite episode length, with each episode decreasing the reward by at least 0.20.
% The nonlinear pendulum system with controller generated by CMA-ES refinement in Table \ref{tab:symbolic} has solutions that generate large negative returns for finite episode length 1000. Specifically, with the semi-implicit Euler scheme, we find transients
%
% 1) at native step size $h=0.05$, with returns as low as $-1210.94$;
%
% 2) at half-native step size $h=0.025$, with returns as low as $-1992.22$;
%
% 3) at quarter-native step size $h=0.0125$, with returns as low as $-4301.13$.
\end{theorem}
%
% \todo[inline]{@Kevin do we have analogous theorem with orbits proven for the 9A CMA finetuned controller}
%

\begin{table*}[h!]
%\captionsetup{font=scriptsize}
\caption{Comparison of different controllers for the pendulum. Mean $\pm$ std.dev. rounded to decimal digit, returns over $100$ episodes reported for different $F_{test}$ (larger the better). $F_{test}=F_{train}$ marked in bold. In this case mean return is equal to negative accumulated penalty. Absolute return discrepancies measure discrepancy in episodic return between different schemes (E/SI) for the same IC (smaller the better). The meaning of observation vector at given time $t$, $x_0 = \cos{\theta(t)}$, $x_1 = \sin{\theta(t)}$, $x_2 = \omega(t) = \theta(t)^\prime$. }
\label{tab:symbolic}
\begin{center}
%\begin{small}
\begin{sc}
\tiny
%\vspace{-.2cm}
\begin{tabular}{cccc|cc|c}
\toprule
 &  & \multicolumn{4}{c}{mean return for given $F_{test}$} & \\
 &  & \multicolumn{2}{c}{$\mathbf{h=0.05}$} & \multicolumn{2}{c}{$h=0.025$} & \multicolumn{1}{c}{ discrepancy}\\
origin & formula & \textbf{SI} & E & SI & E & return E\ /\ SI
\\
\midrule
Alg.~\ref{alg:symbolic}, 3.analytic (symb. $k=9$)& $((((1.30\cdot x_2 + 4.18\cdot x_1)x_0) + 0.36x_1) / -0.52)$& $-207\pm183$ & $-604\pm490$ & $-431\pm396$ & $-910\pm853$ & $479\pm416$\\
Alg.~\ref{alg:symbolic}, 3.CMA-ES (symb. $k=9$)& $((((-10.59x_2 + -42.47x_1)x_0) + 1.2x_1) / 5.06)$ & $-165\pm113$ & $-659\pm461$ & $-331\pm225$ & $-1020\pm801$ & $538\pm401$ \\
Alg.~\ref{alg:symbolic}, small NN & $10$ neurons distilled small NN & $-157\pm99$ & $-304\pm308$ & $-311\pm196$ & $-290\pm169$ & $188\pm285$ \\
\midrule
\cite{symbolic} $(a_1)$ & $-7.08x_1 - (13.39x_1 + 3.12x_2) / x_0 + 0.27$ & $-150\pm87$ & $-703\pm445$ & $-318\pm190$ & $-994\pm777$ & $577\pm401$\\
\midrule
TD3 training & deep NN & $-149\pm86$ & $-138\pm77$ & $-298\pm171$ & $-278\pm156$ & $18\pm38$\\
\bottomrule
\end{tabular}
\end{sc}
%\end{small}
\end{center}
%\vskip -0.1in
\end{table*}
%

%\vspace{-.25cm}

%
\section{Systematic Robustness Study}\label{sec::robustness}
We consider a controller to be {\em robust} when it has ``good" return statistics at the native simulator and step size, which persist when we change simulator and/or decrease step size. If a degradation of return statistics on varying the integrator or step size is identified, we wish to identify the source. 
% That is, we want to identify the object/objects in the dynamics responsible for the poor return.
%
% In this study, we focus on the behaviour of specific numerical simulators under the effect of controllers computed using black box AI methods, rather than the behaviour of the ordinary differential equations themselves. The reasons are twofold.
%
% 1. To solve an ODE plant model coupled with a controller (neural network or otherwise), a numerical integration scheme is needed unless the controller and model are very simple. It is not feasible to study the ODE plant model directly.
%
% 2. For a controller trained with reinforcement learning, the native simulator is an intrinsic part of the control design. We want to assess how robust the control performance is under perturbation of the simulator, which is related to but different from the behaviour of the ODE plant.
%
% \begin{remark}
% To distinguish periodic orbits in the ODEs from those observed using a simulator, we will sometimes refer to the latter as \textit{simulated periodic orbits}. In this study, we are primarily interested in proving non-robustness, rather than proving robustness, hence the focus on the simulator dynamics rather than the ODE dynamics.
% \end{remark}
%
\subsection{Background on Persistent Solutions and Orbits}
\label{sec:persistent}
%Let us start with the setting of our study. 
Consider a MDP tuple $(\sta,\act, F, r, \rho_0, \gamma)$, a precision parameter $\varepsilon > 0$, a policy $\pi\colon\sta\to\act$ (trained using $F_{train}$ and tested using $F_{test}$), a desired equilibrium $\hat{s}$ (corresponding to the maximized reward $r$), and episode length $N$. 
\begin{definition}
\label{def:periodic}
We call a \emph{persistent periodic orbit (PO)} (of period $n$) an infinite MDP trajectory $(s_0,a_0,r_1,s_1,a_1,\dots)$, such that $s_{kn}=s_0$ for some $n>1$ and all $k\in\mathbb{N}$, %, and nonequality otherwise. Moreover, it holds 
and such that $\|\hat{s}-s_j\|>\varepsilon$ for all $j\geq 0$.
\end{definition}
\begin{definition}
\label{def:persistent}
A finite MDP trajectory of episode length $N$ $(s_0,a_0,p_1,s_1,a_1,\dots, s_N)$ such that $\|\hat{s}-s_j\|>\varepsilon$ for all $0\leq j\leq N$ is called a \emph{persistent solution}.
\end{definition}
Locating the objects in dynamics responsible for degradation of the reward is not an easy task, as they may be singular or local minima of a non-convex landscape. For locating such objects we experimented with different strategies, but found the most suitable the evolutionary search of \emph{penalty maximizing solutions}. The solutions identified using such a procedure are necessarily stable. We introduce a measure of 'badness' of persistent solutions and use it as a search criteria.
\begin{definition}
\label{def:persistent_criterion}
We call \emph{a penalty value}, a function %defined over the state and action pairs
$p\colon\sta\times\act\to\R_+$, such that for a persistent solution/orbit the accumulated penalty value is bounded from below by a set threshold $M\gg 0$, that is   $\sum_{i=0}^{N-1}{p(s_i, a_i)}\geq M$.
\end{definition}
\begin{remark}
The choice of particular penalty in Def.~\ref{def:persistent_criterion} depend on the particular studied example. We choose the following penalties in the studied problems.

%\vspace{-.15cm}

1. $p(s,a) = -r(s,a)$ for pendulum.

%\vspace{-.15cm}
2. $p(s,a) = -r(s) + 0.5(\theta^\prime)^2 + 0.5(x^\prime)^2$ for cartpole swing-up. Subtracting from the native reward value $r(s) = \cos{\theta}$ the scaled sum of squared velocities (the cart and pole) and turning off the episode termination condition. This allows capturing orbits that manage to stabilize the pole, but are unstable and %undesirably 
keep the cart moving.
The threshold $M$ in Def.~\ref{def:persistent_criterion} can be set by propagating a number of trajectories with random IC and taking the maximal penalty as $M$.
\end{remark}
\begin{remark}
For a PO, the accumulated penalty admits a linear lower bound, i.e.
% lower bound for linear penalty (Def.~\ref{def:transient}) accumulation rate that holds true for infinite time, 
$\sum_{m=0}^{n-1} p(s_m,a_m) \ge Cn$ for some $C>0$. Thm.~\ref{thm-land} implies $C=0.14$ for the POs in Tab.~\ref{orbit_table_1} in the Appendix.
\end{remark}
\subsection{Searching for and Proving Persistent Orbits}
We designed a pipeline for automated persistent/periodic orbits search together with interval proof certificates. By an interval proof certificate of a PO we mean interval bounds within which a CAP that the orbit exist was carried out applying the Newton scheme (see Sec.~\ref{sec:CAP}), whereas by a proof certificate of a persistent solution (which may be a PO or not) we mean interval bounds for the solution at each step, with a bound for the reward value, showing that it does not stabilize by verifying a lower bound $\|\hat{s}-s_t\|>\varepsilon$. The search procedure is implemented in Python, while the CAP part is in Julia, refer Sec.~\ref{sec:cap_methodology} for further details. 
\begin{algorithm}[h!]
{\small
   \caption{Persistent Solutions/Orbits Search \& Prove}
   \label{alg:transient}
\begin{algorithmic}[1]
   \INPUT $F_{test}$; control policy $\pi$; $h$-parameters of the evolutionary search; penalty function $p$; trajectory length; search domain;
   \OUTPUT interval certificates of persistent/periodic orbits;
   \FOR{\textbf{each} MDP}
       \FOR{ number of searches} 
        \STATE initialize CMA-ES search within specified bounds;
        \STATE search for a candidate maximizing penalty $p$ during the fixed episode length;
        \ENDFOR
        \STATE order found candidates w.r.t. their $p$ value;        
   \ENDFOR
   \FOR{\textbf{each} candidate}
   \STATE search for nearby periodic orbit with Newton's method correction applied to suitable sub-trajectory;
   \IF{potential periodic orbit found}
   \STATE attempt to prove existence of the orbit with Thm.~\ref{theorem-radpol};
   \IF{proof successful}
   \STATE return an interval certificate of the orbit;
   \ELSE
   \STATE return proof failure;
   \ENDIF
   \ELSE
   \STATE return periodic orbit not found;
   \ENDIF
   \STATE produce and return an interval certificate of the uncontrolled solution;
   \ENDFOR
\end{algorithmic}\label{algo_2}}
\end{algorithm}

\subsection{Findings: Pendulum}
\begin{figure}[h!]
\centering
\begin{subfigure}[t]{0.45\columnwidth}
\includegraphics[scale=0.21]{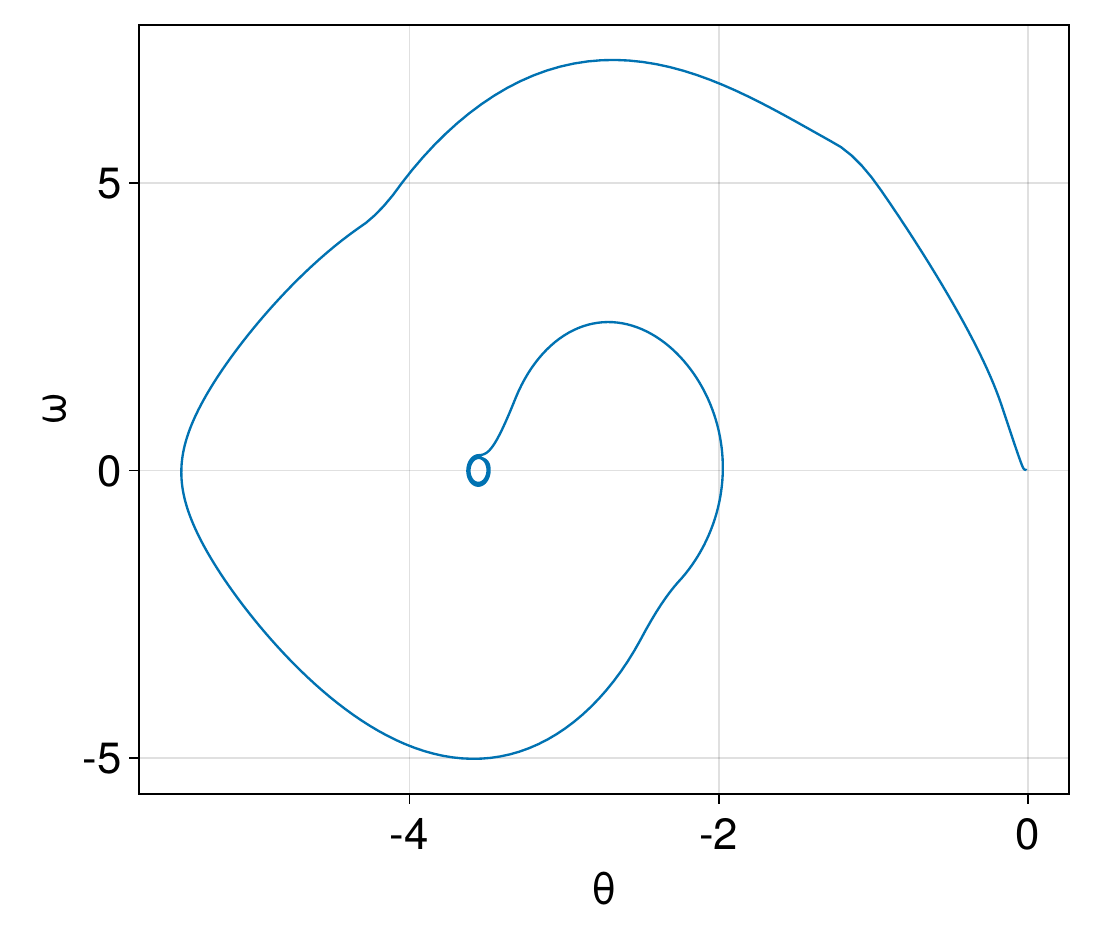}
\end{subfigure}
\begin{subfigure}[t]{0.45\columnwidth}
\includegraphics[scale=0.21]{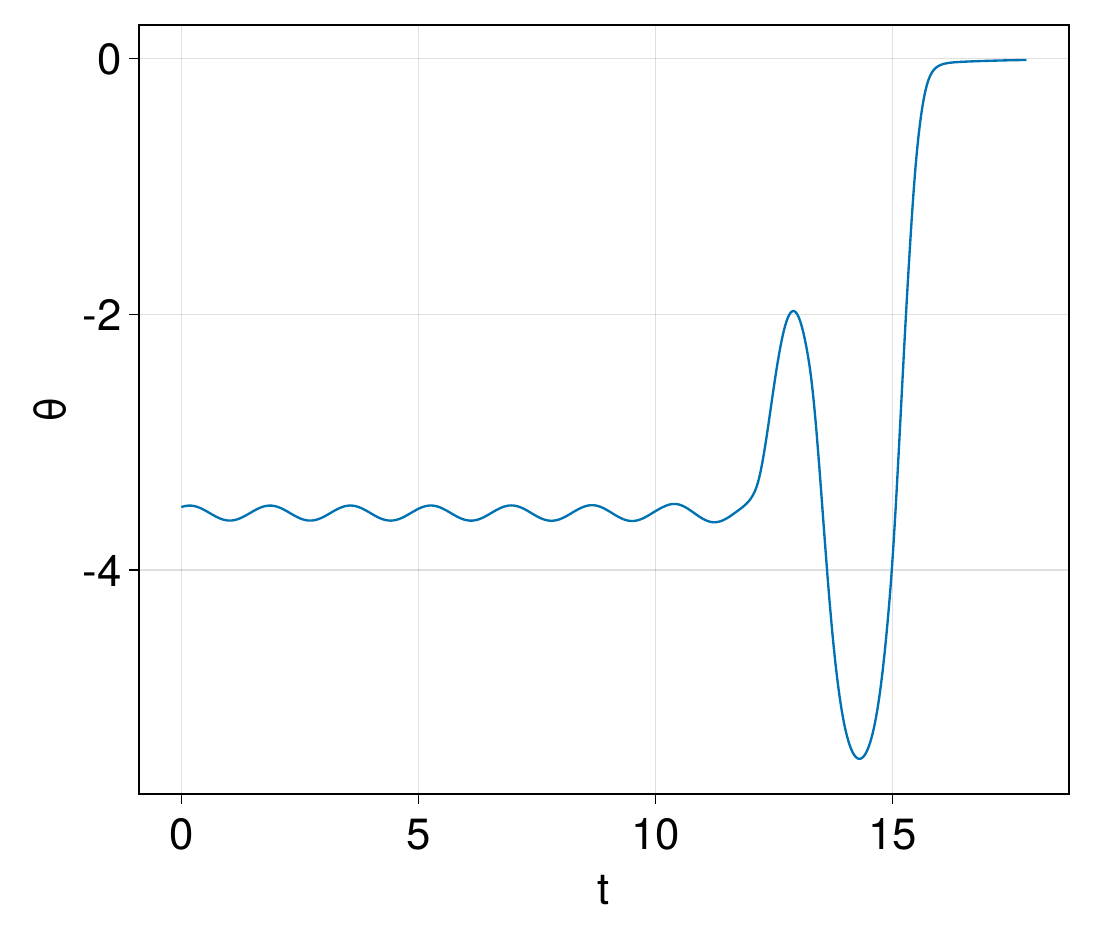}
\end{subfigure}
%\vspace{-.2cm}
\caption{A persistent solution with poor reward $\approx -7527$ over episode length 1000 with step size $h=0.0125$, plotted until near-stabilization at $t=17.825$. Left: plot in phase space. Right: time series of $\theta$. Other data for this solution is in Appendix Tab.\ref{7A_CMA_transients}.}\label{fig:transient2}
\end{figure}

% {\color{blue}We computed mean rewards associated to several feedforward neural networks with various depths, widths and activation functions. Empirical simulation from sampled initial conditions supported the hypothesis that these controllers stabilized the $\theta=0$ state. Strictly speaking, this was not true; the controlled system did not have a steady state at $(\omega,\theta)=(0,0)$, since the controller did not apply zero torque at that state. However, in all cases, the controlled system had a stable equilibrium close to $(0,0)$.}
%
%
%
%
%\subsubsection{Symbolic Controllers}\label{sec:pendulum-symbolic}
% We identified instances in which the symbolic controllers were not robust with respect to simulator or step size. 
% With the training step size and simulator, many of these controllers had better mean returns than the neural networks from which they were derived. 
Changing simulator or step size resulted in substantial mean return loss (see Tab.~\ref{tab:symbolic}), and simulation revealed stable POs (see Fig.~\ref{fig:simulations}). We proved existence of POs using the methods of Section \ref{sec:CAP}--\ref{sec:map}. Proven POs are presented in tables in App.~\ref{sec:proven_orbits}. See also Fig.~\ref{fig:transient2}, where an persistent solution shadows an unstable PO before converging to the stable equilibrium. We present proven persistent solutions in the tables in App. \ref{sec:proven_orbits}.

Comparing the mean returns in Tab.~\ref{tab:symbolic} we immediately see that deep NN controller performance does not deteriorate as much as for the symbolic controller, whereas the small net is located between the two extremes. This observation is confirmed after we run Alg.~\ref{alg:transient} for the symbolic controllers and NN. In particular, we did not identify any stable periodic orbits or especially long persistent solutions. However, the Deep NN controller is not entirely robust, admitting singular persistent solutions achieving returns far from the mean; refer to Tab.~\ref{tab:cartpoletransients}. On the other hand, the small $10$ neuron NN also seems to be considerably more robust than the symbolic controllers. For the case $F_{test}({\rm E},0.05)$ the average returns are two times larger than for the symbolic controllers, but still two times smaller than for the deep NN. However, in the case $F_{test}({\rm E}, 0.05)$, the average returns are close to those of the deep NN contrary to the symbolic controllers. The small NN compares favorably to symbolic controllers in terms of E/SI return discrepancy metrics, still not reaching the level of deep NN. This supports our earlier conjecture (Sec.~\ref{Sec:intro}) that controller robustness is proportional to the parametric complexity.
% {\color{blue} Based on this observation we formulate our corollary that the controller robustness is proportional to the number of parameters, hence one should be going for simple controllers, still having considerable amount of parameters to avoid robustness issues}
%\vspace{-.25cm}
\begin{table}
%\captionsetup{font=scriptsize}
\caption{Examples of persistent solutions found by the persistent solutions Search \& Prove Alg.~\ref{alg:transient} for the pendulum maximizing accumulated penalty, episodes of fixed length $N=1000$. The found persistent solutions were the basis for the persistent orbit/solution proofs presented in App.~\ref{sec:proven_orbits}.}
\label{tab:pendulumtransients}
\begin{center}
\begin{small}
\begin{sc}
{\scriptsize
\begin{tabular}{ccc}
\toprule
controller & MDP & $\sum{r(s,a)}$ \\
\midrule
Alg.~\ref{alg:symbolic} ( $k=9$) & (SI) $h=0.05$ & $-9869.6$ \\
Alg.~\ref{alg:symbolic} ( $k=9$) & (SI) $h=0.025$ & $-1995.7$ \\
\midrule
Alg.~\ref{alg:symbolic} small NN & (SI) $h=0.05$ & $-926.8$ \\
Alg.~\ref{alg:symbolic} small NN & (SI) $h=0.025$ & $-1578.4$ \\
Alg.~\ref{alg:symbolic} small NN & (E) $h=0.05$ & $-747.3$\\
\midrule
\cite{symbolic} $(a_1)$ & (SI) $h=0.05$ & $-873.8$\\
\cite{symbolic} $(a_1)$ & (SI) $h=0.025$ & $-1667.6$\\
\cite{symbolic} $(a_1)$ & (E) $h=0.05$ & $-5391.1$\\
\midrule
deep NN & (SI) $h=0.05$ & $-426.4$\\
deep NN & (SI) $h=0.025$ & $-818.6$ \\
deep NN & (E) $h=0.05$ & $-401.4$\\
\bottomrule
\end{tabular} }
\end{sc}
\end{small}
\end{center}
%\vskip -0.1in
\end{table}
%
%
% \begin{figure}
% \centering
% \includegraphics[scale=0.34,trim={8mm 5mm 0mm 5mm}]{unstable_transient_7CMA_row11.png}
% \caption{A transient solution generating poor reward $\approx -3361$ over episode length 1000 with step size $h=0.0125$. Left: plot in phase space. Right: time series of $\theta$. Other data for this transient is in Table \ref{7A_CMA_transients}.}\label{fig:transient1}
% \end{figure}
%
\subsection{Findings: Cartpole Swing-Up}
\begin{table*}[htbp]
%\captionsetup{font=scriptsize}
\caption{Mean $\pm$ std.dev. reported, rounded to single decimal digits, of returns over $100$ episodes reported for different $F_{test}$ (larger the better). $F_{test}=F_{train}$ marked in bold. Return discrepancies measure discrepancy in episodic return between different schemes (E/SI) for the same IC (smaller the better). The formula for the symbolic controller with $k=21$ is appears in Appendix Tab.\ \ref{table_controllers_cartpole}}
\label{tab:returns_cartpole}
\begin{center}
\begin{small}
\begin{sc}
\tiny
\begin{tabular}{ccc|cc|c}
\toprule
 &  \multicolumn{4}{c}{mean return for given $F_{test}$} & \\
 &  \multicolumn{2}{c}{$\mathbf{h=0.01}$} & \multicolumn{2}{c}{$h=0.005$} & \multicolumn{1}{c}{ discrepancy}\\
origin & SI & \textbf{E} & SI & E & return E\ /\ SI
\\
\midrule
Alg.~\ref{alg:symbolic}, 3.CMA-ES (symb. $k=21$) & $220.2\pm 96.7$ & $334.3\pm 37$ & $474.6\pm 194.3$ & 	$632.2\pm 119.3$ & $121.9\pm 88.9$ \\
Alg.~\ref{alg:symbolic}, small NN ($25$ neurons) & $273.3\pm 128.7$ & $332.9\pm 79.2$ & $585.1\pm 229.1$ & $683.7\pm 103.3$ & $86.6\pm 135.1$\\
TD3 training & $381.2\pm 9.1$ & $382.9\pm 9$ & $760.9\pm 18.4$ & $764.0\pm 18.1$ & $1.7\pm 0.9$ \\
\bottomrule
\end{tabular}
\end{sc}
\end{small}
\end{center}
\vskip -0.1in
\end{table*}
%As for the pendulum, w
We computed the mean return metrics for a representative symbolic controller, a distilled small NN controller and the deep NN, see Tab.~\ref{tab:returns_cartpole}. For the symbolic controller, the average return deteriorates %significantly 
more when changing the simulator's numerical scheme to other than the native ($F_{train}({\rm E}, 0.01)$). Notably, the E/SI discrepancy is an order of magnitude larger than in the case of deep NN. As for the pendulum, the small NN sits between the symbolic and deep NN in terms of the studied metrics. 
We computed the mean accumulated shaped penalty $p(s,a) = -r(s) + 0.5(\theta^\prime)^2 + 0.5(x^\prime)^2$ for the selected controllers in Tab.~\ref{tab:penalties}. The contrast between the deep NN and the symbolic controller is clear, with the small NN being in between those two extremes. The mean penalty is a measure of the prevalence of persistent solutions. However, we emphasize that the Deep NN controller is not entirely robust and also admits singular persistent solutions with bad returns, refer to Tab.~\ref{tab:cartpoletransients}. Rigorously proving the returns for the deep NN was not possible in this case; see Rem.\ \ref{remark-wrapping}.

\begin{figure}[h!]
    \centering
    \captionsetup[subfigure]{justification=centering}
    \begin{subfigure}[t]{0.475\columnwidth}
     \centering
     \includegraphics[width=\textwidth]{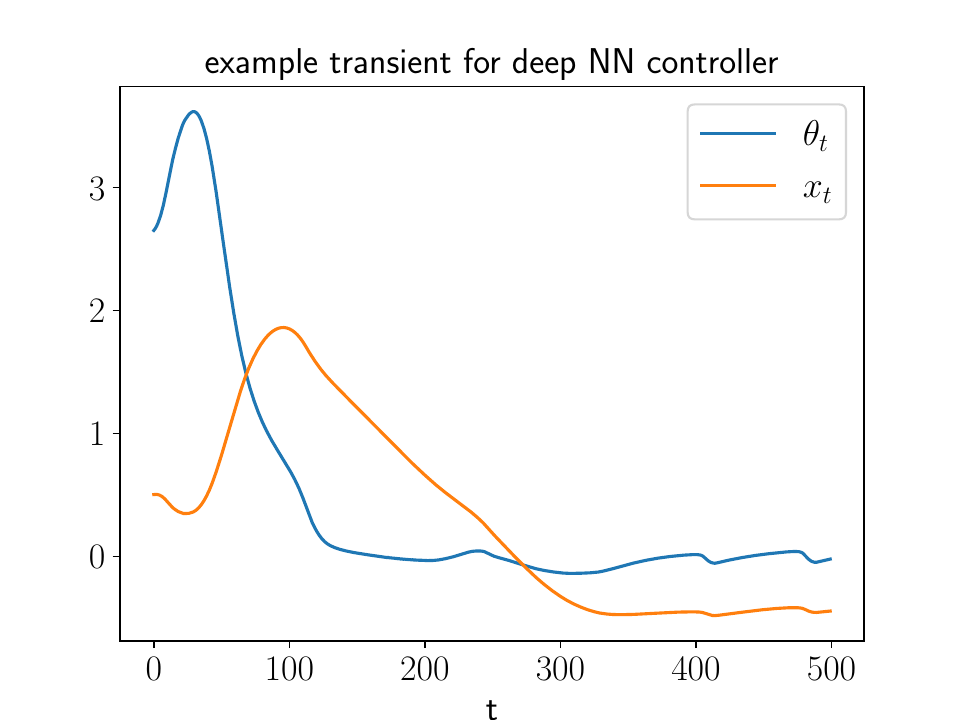}
     \captionsetup{justification=centering}
     \caption{Deep NN controller}
     \label{fig:cartpoletransient1}
    \end{subfigure}
    \hfill
    \begin{subfigure}[t]{0.475\columnwidth}
     \centering
     \includegraphics[width=\textwidth]{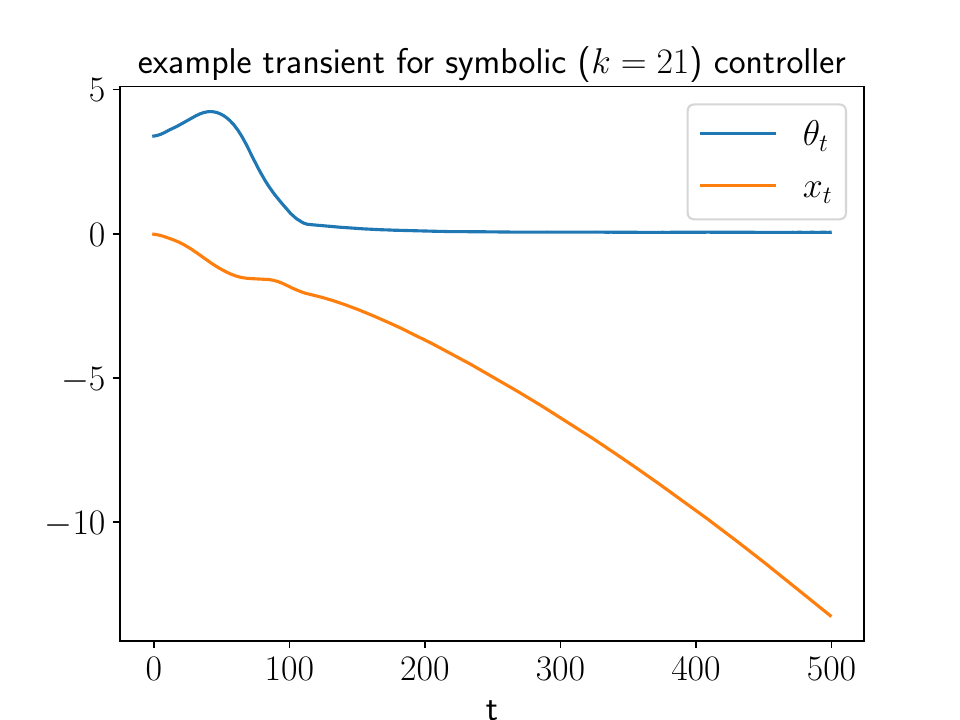}
     \captionsetup{justification=centering}
     \caption{a symbolic controller}
     \label{fig:cartpoletransient2}
    \end{subfigure}
    %\vspace{-.2cm}
    \caption{The persistent solutions (evolution of $(\theta,x)$ (Def.~\ref{def:persistent}) for cartpole swing-up problem found with Alg.~\ref{alg:transient} that maximize accumulated penalty $\sum p(s,a) = \sum{-r(s) + 0.5(\theta^\prime)^2 + 0.5(x^\prime)^2}$ over episodes of length $2000$ without terminations, using SI with $h=0.01$. (a) $\sum{p(s,a)} = -306$; (b) $\sum{p(s,a)} = 37746$. }
    %\vspace{-10pt}
    \label{fig:cartpole_transients}
\end{figure}

Investigating the persistent solutions found with Alg.~\ref{alg:transient} in Fig.~\ref{fig:cartpole_transients} we see that in case $F_{test}({\rm SI}, 0.01)$ the symbolic controller admits bad persistent solutions with $x_t$ decreasing super-linearly, whereas $\theta$ stabilizes at $\theta\sim0.01$. In contrast, the deep NN exhibits fairly stable control with small magnitude oscillations. This example emphasizes the shaped penalty's usefulness in detecting such bad persistent solutions. We can see several orders of magnitude differences in the accumulated penalty value for the deep NN controller vs. the symbolic controller case. We identify and rigorously prove an abundance of persistent solutions for each of the studied symbolic controllers. For example, we can prove:

\begin{theorem}
For the symbolic controller with complexity $k=21$ and native step size $h=0.01$, there are 2000-epsiode persistent solutions of the cartpole swing-up model with accumulated penalty $\geq 2.66\times 10^5$ for the explicit scheme, and $\geq 3.77\times 10^5$ for the semi-implicit scheme. With the Small NN controller, the conclusions hold with accumulated penalties $\geq 6263$ and $\geq 2.68\times 10^6$.
\end{theorem}

\begin{table}
%\captionsetup{font=scriptsize}
\caption{Examples of persistent solutions found by the transient solutions Search \& Prove Alg.~\ref{alg:transient} for the cartpole-swingup maximizing the accumulated penalty, episodes of fixed length $N=2000$ without taking into account the termination condition. The found persistent solutions were the basis for the persistent orbit/solution proofs presented in App.~\ref{sec:proven_orbits_cartpole}}
\label{tab:cartpoletransients}
\begin{center}
\begin{small}
\begin{sc}
{\scriptsize
\begin{tabular}{ccc}
\toprule
controller & MDP & $\sum{r(s,a)}$ \\
\midrule
Alg.~\ref{alg:symbolic} ( $k=21$) & (SI) $h=0.01$ & $-41447.2$ \\
Alg.~\ref{alg:symbolic} ( $k=21$) & (SI) $h=0.005$ & $-11204.3$ \\
Alg.~\ref{alg:symbolic} ( $k=21$) & (E) $h=0.01$ & $-29878.0$ \\
Alg.~\ref{alg:symbolic} ( $k=21$) & (E) $h=0.005$ & $-8694.3$ \\
\midrule
Alg.~\ref{alg:symbolic} small NN & (SI) $h=0.01$ & $-2684696.8$ \\
Alg.~\ref{alg:symbolic} small NN & (SI) $h=0.005$ & $-798442.3$ \\
Alg.~\ref{alg:symbolic} small NN & (E) $h=0.01$ & $-520.9$\\
Alg.~\ref{alg:symbolic} small NN & (E) $h=0.005$ & $-2343.8$\\
\midrule
deep NN & (SI) $h=0.01$ & $306.6$\\
deep NN & (SI) $h=0.005$ & $-396074.9$ \\
deep NN & (E) $h=0.01$ & $226.5$\\
deep NN & (E) $h=0.005$ & $-1181.7$\\
\bottomrule
\end{tabular}}
\end{sc}
\end{small}
\end{center}
%\vskip -0.1in
\end{table}
We demonstrate persistent solutions for each considered controller in Tab.~\ref{tab:cartpoletransients}. The found persistent solutions were the basis for the persistent orbit/solution proofs presented in App.~\ref{sec:proven_orbits_cartpole}. The symbolic and small NN controllers admit much worse solutions with increasing velocity, as illustrated in Fig.~\ref{fig:cartpoletransient2}. Deep NN controllers admit such bad solutions when tested using smaller time steps ($(E,0.005), (SI,0.005)$); see examples in Tab.~\ref{tab:cartpoletransients}. They also exhibit persistent periodic solutions, albeit with a small $\epsilon$; see Fig.~\ref{fig:cartpoletransient1}. We have proven the following.

\begin{theorem}
For $h$ close to\footnote{The exact step size is smaller than $h$, with relative error up to 2\%. See App.~\ref{sec:proven_orbits_cartpole} for precise values and detailed data for the POs.} $0.005$ and $h=0.01$ (native), the cartpole swing-up model has POs for (E) and (SI) with the deep NN controller. The mean penalties along orbits are greater than $-0.914$ and are persistent\footnote{With respect to the translation-invariant seminorm $||(x,\dot x,\theta,\dot\theta)|| = \max\{|\dot x|,|\theta|,|\dot\theta|\}$} with $\epsilon\geq 0.036$.
\end{theorem}
\begin{remark}\label{remark-wrapping}
We were not able to rigorously compute the penalty values of the persistent solutions for the deep NN controller due to wrapping effect of interval arithmetic calculations \cite{Tucker2011}, which is made much worse by the width of the network (400,300) and the long epsiode length (which introduces further composition). However, this is not a problem for the periodic orbits: we enclose them using Theorem \ref{theorem-radpol}, which reduces the wrapping effect. 
\end{remark}
\begin{table}
%\captionsetup{font=scriptsize}
\caption{Comparison of different controllers for the cartpole swing-up for $h=0.01$.
Mean and std.dev. (after $\pm$) reported of accumulated penalties $\sum p(s_k) = \sum -r(s_k) + 0.5(\theta_k^\prime)^2 + 0.5(x_k^\prime)^2$ (larger the worse) over $100$ episodes reported for different $F_{test}$.  $F_{test}=F_{train}$ marked in bold. Controllers same as in Tab.~\ref{tab:returns_cartpole}. }
\label{tab:penalties}
\begin{center}
\begin{small}
\begin{sc}
\scriptsize
%\vspace{-.4cm}

\begin{tabular}{cccc}
\toprule
% &  \multicolumn{2}{c}{$\mathbf{h=0.01}$}\\
origin & SI & \textbf{E}\\
\midrule
Alg.~\ref{alg:symbolic}, 3.CMA-ES (symb. $k=21$) & $3123.0\pm 719.9$ & $2257.2\pm 234.1$ \\
Alg.~\ref{alg:symbolic}, small NN ($25$ neurons) & $1413.4\pm 9670.1$ & $404.2\pm 148.4$\\
TD3 training & $335.7\pm 64.7$ & $425.6\pm 72.1$ \\
\bottomrule
\end{tabular}
\end{sc}
\end{small}
\end{center}
%\vskip -0.1in
\end{table}

\section{Codebase}
Our full codebase is written in Python and Julia shared in a github repository \cite{code}. The reason why the second part of our codebase is written in Julia is the lack of a suitable interval arithmetic library in Python. 
The Python part of the codebase consists of four independent parts -- scripts: deep NN policy training, symbolic/small NN controller regression, regressed controller fine-tuning and periodic orbit/persistent solution searcher. All controllers that we use are implemented in Pytorch \cite{pytorch}. For the deep NN policy training we just use the Stable-baselines 3 library \cite{sb3}, which outputs a trained policy (which achieved the best return during training) and the training replay buffer of data. For the symbolic regression we employ the PySR lib. \cite{pysr}. For the regressed controller fine-tuning we employ the pycma CMA-ES implementation \cite{pycma}. 
Our implementation in Julia uses two external packages: IntervalArithmetic.jl \cite{david_p_sanders_2022_7257716} (for interval arithmetic) and ForwardDiff.jl \cite{RevelsLubinPapamarkou2016} (for forward-mode automatic differentiation). These packages are used together to perform the necessary calculations for the CAPs. 
\section{Conclusion and Future Work}
Our work is a first step towards a comprehensive robustness study of deep NN controllers and their symbolic abstractions, which are desirable for deployment and trustfulness reasons. Studying the controllers' performance in a simple benchmark, we identify and prove existence of an abundance of persistent solutions and periodic orbits. Persistent solutions are undesirable and can be exploited by an adversary. Future work will apply the developed methods to study higher dimensional problems often used as benchmarks for continuous control.
\section{Acknowledgements}
The project is financed by the Polish National Agency for Academic Exchange. The first author has been supported by the Polish National Agency for Academic Exchange Polish Returns grant no. PPN/PPO/2018/1/00029 and the University of Warsaw IDUB New Ideas grant. This research was supported in part by PL-Grid Infrastructure. 
% In the unusual situation where you want a paper to appear in the
% references without citing it in the main text, use \nocite
\newpage
\nocite{langley00}

\bibliography{aicap}

\newpage
\appendix
\onecolumn
\section{Studied Dynamical Systems}
In this section we describe the details of studied problems. The continuous dynamical system with their discretizations.
\subsection{Pendulum}
\label{sec:pendulum}
\subsubsection{Continuous System}
We study numerical discretizations of the following continuous dynamical system governing the motion of a simple pendulum
\[
    \theta^{\prime\prime}(t) = \frac{3u(t)}{l^2m} + \frac{3g\sin(\theta(t))}{2l},
\]
where $\theta(t)$ is the current angle at time $t$ of the pendulum, and $u(t)$ is the controller input, $l$, $m$, $g$ are the pendulum length, mass and the gravitational constant respectively. We set them to the values used in the environment code, i.e., $l=1, m=1, g=10$. The uncontrolled model admits the unstable equilibrium at $\theta=0$, and the stable equilibrium at $\theta=\pm \pi$.

Let us introduce auxiliary variable $\omega = \theta^\prime$, and the following extended system
\begin{subequations}
\label{eq:extended}
\begin{align}
    \theta^\prime(t) &= \omega(t),\\
    \omega^{\prime}(t) &= \frac{3u(t)}{l^2m} + \frac{3g\sin(\theta(t))}{2l},
\end{align}
\end{subequations}
let us denote $x(t) = [\theta(t), \omega(t)]$, and by $f(x(t))$ we denote the right hand side of \eqref{eq:extended}.

\subsubsection{Discrete Systems}
\label{sec:pendulumdiscrete}
In practice, we will study various discrete dynamical systems arising from discretizing \eqref{eq:extended} with timesteps $t_0,t_1,\dots,t_k$ and fixed constants $l$, $m$, $g$, and employing various numerical methods mentioned below.

The particular numerical methods that we study include
\paragraph{Explicit Euler, fixed time-step}
\begin{align*}
    t_k &= t_{k-1} + h,\\
    \bar{\theta}_k &= \bar{\theta}_{k-1} + h\bar{\omega}_{k-1}(t),\\
    \bar{\omega}_k &= \lfloor\bar{\omega}_{k-1} + hf^2(x_{k-1}, \lfloor u_{k-1}\rfloor)\rfloor,
\end{align*}
we denote the formula for $x_k=(\bar{\theta}_k,\bar{\omega}_k)$ by $g_{ee}(t_{k-1}, x_{k-1})$. Where $h$ is the uniform fixed time-step. We denote $\lfloor u_{k-1}\rfloor$, and $\lfloor\bar{\omega}_k \rfloor$ denote the clipped values, i.e. if $u_{k-1}$ exceeds the range $[-2,2]$ its value is clipped to the closest value in the range (whereas the velocity is clipped to the range $[-8,8]$).
\paragraph{Semi-implicit Euler} This is the original numerical method used for solving the pendulum dynamics in the OpenAI gym package \cite{gym} (with $h=0.05$)
\begin{align*}
    t_k &= t_{k-1} + h,\\
    \bar{\omega}_k &= \lfloor\bar{\omega}_{k-1} + hf^2(x_{k-1}, \lfloor u_{k-1}\rfloor)\rfloor,\\
    \bar{\theta}_k &= \bar{\theta}_{k-1} + h\bar{\omega}_k,
\end{align*}
where $x_k=(\bar{\theta}_k, \bar{\omega}_k)$, and we denote the formula for $x_k$ by $g_{sie}(t_{k-1}, x_{k-1})$.
We call the method semi-explicit, however it is cooked up for the particular case of the pendulum. In the first step the new velocity is computed and then in the second step the angle is updated using the new velocity.
%
%
%
% \paragraph{Implicit Euler, fixed time-step}
% \begin{align*}
%     t_k &= t_{k-1} + h,\\
%     x_k &= x_{k-1} + hf(x_{k}, \lfloor u_{k}\rfloor),
% \end{align*}
% where $h$ is the uniform fixed time-step. We solve for the implicitly defined $x_k$ using an iterative process guaranteed to converge for a sufficiently small time-step. Denote the resulting output $x_k$ by $g_{ie}(t_{k-1}, x_{k}, \lfloor u_{k-1}\rfloor)$.
%%%%%%%%%%%%%%%%%%%%%%%%%%%%%%%%%%%%%%%%%%%%%%%%%%%%%%%%%%%%%%%%%%%%%%%%%%%%%%%
%%%%%%%%%%%%%%%%%%%%%%%%%%%%%%%%%%%%%%%%%%%%%%%%%%%%%%%%%%%%%%%%%%%%%%%%%%%%%%%
\section{Cartpole Swing-up}
\label{sec:cartpole}
\subsection{Continuous System}
We used the implementation \cite{cartpolecode}, which is slightly modified version used in \cite{hacartpole}. The motion of the cartpole is determined by the following dynamical system
\begin{align*}
\theta^{\prime\prime} &= \left(-3(m_p+l)(\theta^\prime)^2\sin{\theta}\cos{\theta} + 6(m_p+m_c)g\sin{\theta} + 6(u-fx^\prime)\cos{\theta}\right) / (4(m_p+m_c)l-3(m_p+l)\cos^2{\theta}) = f_1(\theta, \theta^\prime, x^\prime),\\
x^{\prime\prime} &= \left(-2(m_p+l)\cdot(\theta^{\prime})^2\cdot\sin{\theta} + 3m_pg\sin{\theta}\cos{\theta}+4u-4fx^\prime\right) / (4(m_p+m_c)-3m_p\cos^2{\theta}) = f_2(\theta, \theta^\prime, x^\prime, u),
\end{align*}
where $u$ is the control input at given time (input from range $[-1,1]$ is multiplied by $10$). We set the constants as in the original environment code, pole mass $m_p=0.5$, pole length $l = 0.6$, cart mass $m_c=0.5$, gravity const. $g=9.82$, fricition const. $f=0.1$
\subsection{Discrete Systems}
The continuous dynamical system above is discretized using two numerical schemes that we present below.
\subsubsection{Explicit Euler} This is the numerical method applied in the original implementation (with $h=0.01$).
\begin{align*}
t_k &= t_{k-1} + h,\\
x_k &= x_{k-1} + hx^\prime_{k-1},\\
\theta_k &= \theta_{k-1} + h\theta^\prime_{k-1},\\
x^\prime_k &=x^\prime_{k-1} + h\cdot f_2\left(\theta_{k-1}, \theta^\prime_{k-1}, x^\prime_{k-1}, 10\cdot \lfloor u_{k-1}\rfloor \right),\\
\theta^\prime_k &= \theta_{k-1}^\prime + h\cdot f_1\left( \theta_{k-1},\theta^\prime_{k-1},x^\prime_{k-1}\right),
\end{align*}
where $\lfloor\cdot\rfloor$ is clipping the input to interval $[-1,1]$.
\subsubsection{Semi-implicit Euler} This is another numerical method that we implemented for our robustness study, based on the method used in pendulum (with $h=0.01$).
\begin{align*}
t_k &= t_{k-1} + h,\\
x^\prime_k &=x^\prime_{k-1} + h\cdot f_2\left(\theta_{k-1}, \theta^\prime_{k-1}, x^\prime_{k-1}, 10\cdot \lfloor u_{k-1}\rfloor \right),\\
\theta^\prime_k &= \theta_{k-1}^\prime + h\cdot f_1\left( \theta_{k-1},\theta^\prime_{k-1},x^\prime_{k-1}\right),\\
x_k &= x_{k-1} + hx^\prime_{k},\\
\theta_k &= \theta_{k-1} + h\theta^\prime_{k}.
\end{align*}
By the semi-implicit scheme in this case we mean that velocities $x_k^\prime,\theta_k^\prime$ are first updated using the accelerations from the previous step (depending on $\theta_{k-1},\theta^\prime_{k-1},x^\prime_{k-1}$), then positions $x_k,\theta_k$ are updated using the current velocities $x^\prime_k,\theta^\prime_k$.
\section{TD3 RL agents training curves}
\label{sec:training}
\begin{figure}[h!]
    \centering
    \captionsetup[subfigure]{justification=centering}
    \begin{subfigure}[t]{0.4\columnwidth}
     \centering
     \includegraphics[width=\textwidth]{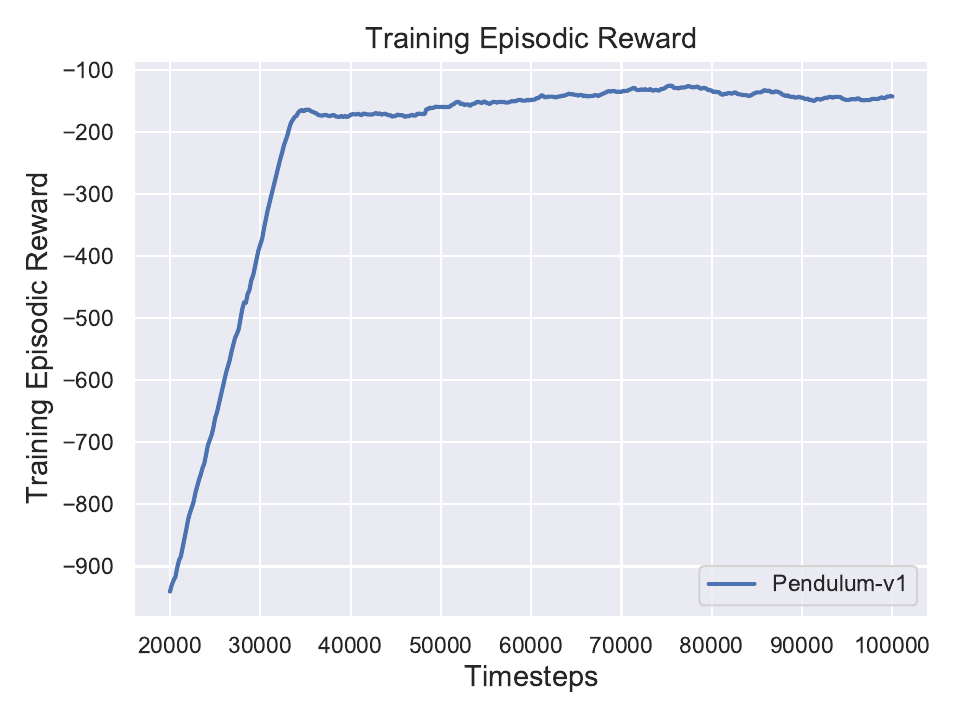}     \captionsetup{justification=centering}
     \caption{Pendulum}
     \label{fig:train_pendulum}
    \end{subfigure}
    \hfill
    \begin{subfigure}[t]{0.4\columnwidth}
     \centering
     \includegraphics[width=\textwidth]{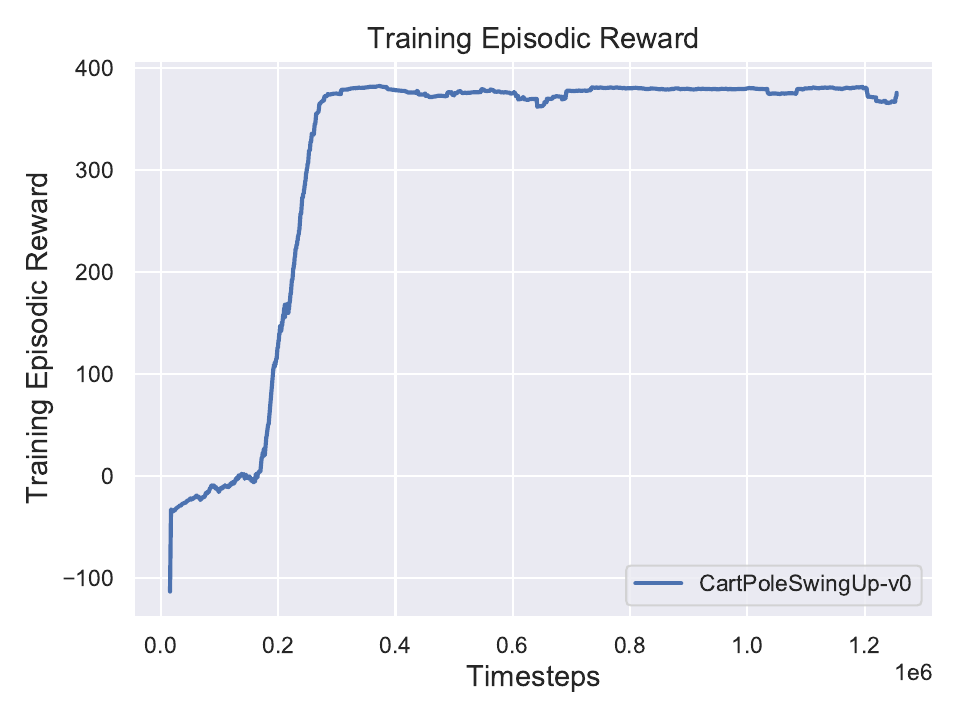}
     \captionsetup{justification=centering}
     \caption{Cartpole Swing-up}
     \label{fig:train_cartpole}
    \end{subfigure}
    \vspace{-.2cm}
    \caption{Obtained TD3 training curves using the SB3 implementation \cite{sb3}. As the 'deep NN' controller we pick the best according to the episodic return checkpoint obtained within the number of episodes shown.}
    %\vspace{-10pt}
    \label{fig:trainings}
\end{figure}

We report on the RL training of the deep NN controllers studied in this work utilizing the TD3 \cite{td3} algorithm. The resulting plots showing training episodic returns are presented in Fig.~\ref{fig:trainings}.

\section{Rigorous proof methodology: further details}
Here we provide additional details concerning the maps $G$ required for computer-assisted proofs, and the implementation.

\subsection{The map $G$ for the variable step size case}\label{sec:variable-step}
Suppose we want to treat $h$ as variable, rather than an \textit{a priori} fixed constant. Let $\eta:\mathbb{R}^p\rightarrow\mathbb{R}$ be given, and define $G_2:\mathbb{R}^{pm+1}\rightarrow\mathbb{R}^{pm+1}$ by 
\[{
G_2(X) = \begin{pmatrix}
\eta(x_0)
\\
x_0-g(h,x_{m})+(j2\pi,\mathbf{0})
\\
x_1-g(h,x_{0})
\\
x_2-g(h,x_{1})
\\
\vdots
\\
x_m-g(h,x_{m-1})
\end{pmatrix},}
\]
where $X=(h,x_0,\dots,x_m)$. $\eta$ compensates for the addition of another variable. In practice, we choose $\eta$ to be a linear function. 
% This formulation is used when we find an approximate orbit, but Newton's method fails to converge with the map $F_1$. See Appendix \ref{sec:discretized_qpo} for discussion.

\subsection{Implementation details}\label{sec:Julia_implemtation_details}
In our robustness study, the maps $G$ (e.g.\ $G_1$, $G_2$) used for computer-assisted proofs are not globally continuously differentiable. Indeed, we have 1) lack of smoothness of the symbolic controller representation (e.g.\ divisions by zero at some inputs, non-smooth ReLU activation function), and 2) discrete logic rules (e.g.\ clipping, piecewise-linear saturation) in the simulator and/or controller. This makes it difficult to verify that $G:U\rightarrow\mathbb{R}^n$ is locally (i.e.\ in $U$) continuously differentiable. We overcome this with a clever implementation-level trick: we implement $g$ (the discrete-time system defining $G_1$ and $G_2$) in such a way that Julia will return an error if $g$ is evaluated on an interval (or interval vector) input that contains a point where the function is either undefined or non-smooth. 
% Since $F_1$ and $F_2$ are defined in terms of $g$, 
This ensures that successfully evaluating $G_1$ or $G_2$ in Julia on an \textit{interval} automatically proves it is smooth there.

Let us go over how Theorem \ref{theorem-radpol} is verified. First, we identify a candidate for a periodic orbit as in Section \ref{sec::robustness}, denoted $\overline x$ and stored as a vector such that $G(\overline x)\approx 0$.
%This involves processing the data such that we obtain a vector $\overline x$ with $F(\overline x)\approx 0$, and applying Newton's method to improve the quality of $\overline x$, so that $||F(\overline x)||$ is machine-precision small. 
We use automatic differentiation to calculate $DG(\overline x)$, and then calculate a machine inverse $A$. To get $Y$, we simply calculate $||AG(\overline x)||$ using interval arithmetic, and let $Y$ be the interval supremum. We do the same thing for the bound $Z_0$. These choices result in \eqref{Y-bound} and \eqref{Z0-bound} being satisfied. Note that if $Z_0+Z_2<1$, then $||I-ADG(\overline x)||<1$ and it follows that $A$ must be full rank. To calculate $Z_2$, we use automatic differentiation to first calculate the Jacobian of $G$ at the interval representation of the closed ball (interval vector) $[\overline x]_{r^*} = \{x\in\mathbb{R}^n : ||x-\overline x||\leq r^*\}.$ Note that $r^*$ must be specified beforehand\footnote{We typically start with $r^*=10^{-4}$ and decrease if necessary. It can be necessary to decrease $r^*$ if $Z_2$ is too large for the proof to succeed or $G$ is non-smooth on the candidate domain.}. We compute $Z_2 = \sup||A(DG([\overline x]_{r^*})-DG(\overline x))||,$ where $\sup$ denotes the interval supremum. 
% By the inclusion principle of interval arithmetic, 
This choice of $Z_2$ results in \eqref{Z2-bound} being satisfied. We then compute $Y/(1-Z_0-Z_2)$ and let $r$ be the result of rounding up to the next float. Finally, we check that $Y+r(Z_0+Z_2-1)<0$ and $r\leq r^*$, as desired. 

To contrast, proofs of persistent solutions do not require Theorem \ref{theorem-radpol}. Instead, we reliably simulate persistent solutions by running the simulator initialized at a thin interval IC. In other words, we use interval arithmetic to rigorously track rounding errors caused by the numerical simulator. The amount of steps we can reliably simulate in this way is dependent on the precision of the number system.

\section{Persistence of periodic orbits under discretization}\label{sec:discretized_qpo}
For a simple example that can be studied analytically, consider the harmonic oscillator $\ddot x=-x$. Transforming to polar coordinates via $x=r\cos\theta$ and $\dot x=r\sin\theta$, we get the two-dimensional ODE system $$\dot r=0,\hspace{2mm}\dot\theta=1.$$
A periodic orbit corresponds to rotation, that is, $\theta\mapsto\theta+2\pi$. However, with the forward Euler integrator at step size $h$, step $k$ has coordinates $(r_k,\theta_k)=(r_0,\theta_0+kh)$. Unless $h$ and $\pi$ are commensurate, a periodic orbit can not exist. In this case, the set $\{\theta_k \mod 2\pi,\hspace{1mm}k=1,2,\dots\}$ densely fills the interval $[0,2\pi]$. In the topological dynamics sense, this indicates that periodic orbits could persist under discretization as orbits equivalent to an irrational rotation of the circle $\mathbb{R}/\mathbb{Z}$. If $h$ and $\pi$ are commensurate -- say, $am=b2\pi$ for some integers $m$ and $b$ -- then $\theta_m = \theta_0 + b2\pi$. Therefore, there is a $m$-step orbit. However, it could be that $m$ is extremely large. 

% An explicit mechanism for the persistence of periodic orbits can be understood through Neimark-Sacker bifurcations as the step size $h$ is varied. Consider the ``non-rotating" simulated periodic orbit with controller 3 and the semi-implicit Euler scheme from \textcolor{red}{finalized\_file\_name\_Symbolic\_simple2\_finetune\_results.csv}, proven with non-fixed step size near $h=0.04836$. Let $F_h:\mathbb{R}^2\rightarrow\mathbb{R}^2$ denote the composition of the semi-implicit integrator with itself $m$ times, for step size variable $h$. For some $h$ near $0.04836$, our proof guarantees that for some $(\theta_0,\omega_0)\approx(-2.7296,-0.1982)$, we have a fixed point: $F_h(\theta_0,\omega_0)=(\theta_0,\omega_0)$. One can show that the Jacobian $DF_h(\theta_0,\omega_0)$ is non-singular, so in particular this conclusion remains robust for a continuous range of $h$ near $0.04836$. However, as $h$ increases from $0.04836$ to $0.05$, a Neimark-Sacker bifurcation occurs, characterized by the appearance of pair of complex-conjugate eigenvalues of $DF_h(\theta_0,\omega_0)$ that have absolute value 1. The result is that the fixed point of $F_h$ destabilizes and a stable invariant circle appears. This invariant circle is does not correspond to a simulated periodic orbit for any number of steps, but it traces an oscillatory path through phase space. See Figure \ref{fig:NSB}.

To summarize, a periodic orbit in an ODE could persist as an $m$-step orbit for a possibly large $m$, or it could be equivalent to an irrational circle rotation. The example we saw is artificial, since its periodic orbits (for the ODE) come in continuous families parameterized by the radius in polar coordinates. However, the idea demonstrates why it might not be possible to (easily) prove a periodic orbit for a given step size $h$ and number of steps $m$, despite the appearance of a good numerical candidate.

\section{Proven periodic orbits and persistent solutions for the inverted pendulum}\label{sec:proven_orbits}
In the following pages, we catalogue the periodic orbits and persistent solutions we have proven for the inverted pendulum model. To improve readability, all initial angles, angular velocities and step sizes are truncated to five decimal places. In all cases, numbered controllers reference those in Table \ref{table_controllers_pendulum}. For the periodic orbits (Table \ref{7A-orbit-table} through Table \ref{19A-CMA-pendulum-orbits}), we indicate if the step size is proven exactly (i.e.\ the map $G_1$ is used) or not. The direction column indicates if the orbit rotates counter-clockwise (direction = $+$) or clockwise (direction = $-$). For persistent solutions (Table \ref{7A-persistent-pendulum} through Table \ref{19A-persistent-pendulum}), we integrate forward from ICs $(\theta_0,\omega_0)$ with the specified integrator. In the persistent solution tables, the time $T_p$ denotes one of the following:
\begin{itemize}
\item the first time where the solution satisfies $(\tilde\theta ,\omega)\in [0,10^{-2}]\times[-10^{-2},10^{-2}]$ for $\tilde\theta=\arccos(\cos(\theta))$, or
\item if this does not occur within 1000 steps (episodes), then we let $T_p=1000h$.
\end{itemize}
Note that the step size $h$ of the integrator and the final time $T_p$ are related by the relationship $T_p = mh$, where $m$ is the number of simulation steps. The rewards and returns stated are midpoints of interval-value rewards, of which the latter are guaranteed to enclose the true value. The radii of these intervals are in all cases small, typically around $10^{-14}$.

\begin{table}
\centering
\begin{tabular}{|c|c|c|c|c|c|c|}
\hline
     Numerical Method & $h$ & $m$ & $\theta_0$ & $\omega_0$ & Max Reward   \\
     \hline
Explicit&0.05&28&3.94871&8.0&-0.64228\\Explicit&0.025&55&4.10685&7.83862&-0.68942\\Explicit&0.01&166&0.69262&1.59285&-0.33452\\Explicit&0.005&358&0.69672&1.42118&-0.26396\\Explicit&0.0025&721&0.69597&1.40667&-0.25791\\Explicit&0.001&1839&0.29451&1.00288&-0.24372\\Semi-Implicit&0.01&202&0.20564&1.02174&-0.19888\\Semi-Implicit&0.005&398&0.69922&1.23635&-0.20352\\Semi-Implicit&0.0025&760&0.6974&1.30669&-0.22475\\Semi-Implicit&0.001&1870&0.48466&0.89362&-0.23343\\
      \hline
\end{tabular}
\caption{Summary data for the periodic orbits proven for the inverted pendulum model with the Landajuela et.\ al controller $a_1 = -7.08s_2 - (13.39s_2 + 3.12s_3) / s_1 + 0.27$. All orbits complete a single counter-clockwise rotation.}\label{orbit_table_1}
\end{table}

\begin{table}
\centering
\begin{tabular}{|c|l|}
\hline
Controller & Formula\\
\hline
7A AG& $-((1.074\cdot(x_2 \cdot x_0) + 3.064\cdot x_1 ) / 0.482)$\\
9A AG& $-((((1.303\cdot x_2 + 4.180\cdot x_1) \cdot  x_0) + 0.364\cdot x_1) / 0.519)$\\
13A AG& $ (((x_2\cdot 1.168 + x_1\cdot 4.4618) \cdot  x_0) / ((x_2 \cdot  (-x_2 \cdot  0.014))  -0.207))$\\
17A AG& $(((0.567\cdot x_2 + 2.032\cdot x_1) \cdot  x_0 \cdot  1.381) / ((x_2 \cdot  ((x_2 \cdot  (x_0 \cdot  x_0)) \cdot  -0.034)  -0.112))$\\
19A AG& $(((1.627\cdot x_2 + (x_1 / 0.161)) \cdot  x_0) / ((((x_1 / 0.168) + 0.993\cdot x_2) \cdot  (-x_2 \cdot  0.085))  -0.754))$\\
7A CMA & $ -((2.865\cdot (x_2 \cdot  x_0) + 6.973\cdot x_1 ) / 1.048)$\\
9A CMA & $ ((((-105.902\cdot x_2  -424.711\cdot x_1) \cdot  x_0) + 12.033\cdot x_1) / 50.577)$\\
13A CMA & $ (((x_2\cdot 31.252 + x_1\cdot 122.785) \cdot  x_0) / ((x_2 \cdot  (-x_2 \cdot  1.426))  -11.029))$\\
17A CMA & $ (((4.813\cdot x_2 + 11.061\cdot x_1) \cdot  x_0 \cdot  20.311) / ((x_2 \cdot  (-(x_2 \cdot  (x_0 \cdot  x_0)) \cdot 9.437))  -15.478))$\\
% 19A CMA & $ (((8.138\cdot x_2 + (x_1 / 0.057)) \cdot  x_0) / ((((x_1 / 1.476)  -0.415\cdot x_2) \cdot  (x_2 \cdot  0.601))  -0.492)) $\\
19A CMA & $(((7.943\cdot x_2 + (x_1 / 0.070)) \cdot x_0) / ((((x_1 / 1.567)  -0.335\cdot x_2) \cdot (x_2 \cdot 0.540))  -0.639))$ \\
\hline
\end{tabular}
\caption{Controllers dictionary for interved pendulum. AG refers to controllers refined by analytic gradient, CMA refers to those refined by CMA-ES. For readability, all parameters are truncated to three decimal places and we have distributed negative signs where possible. Note that $x_0=\cos(\theta)$, $x_1=\sin(\theta)$ and $x_2=\dot\theta$.}\label{table_controllers_pendulum}
\end{table}

\begin{table}
\centering
\begin{tabular}{|c|c|c|c|c|c|c|c|c|}
\hline
Controller & Numerical Method & $h$ & $m$ & $\theta_0$ & $\omega_0$ & Direction & Exact $h$? & Max Reward  \\
\hline
7A AG&Explicit&0.05&23&5.11631&6.8593&+&No&-1.47086\\7A AG&Explicit&0.05&23&13.73324&-6.8593&-&No&-1.47086\\7A AG&Explicit&0.025&46&6.05378&4.24904&+&Yes&-1.25634\\7A AG&Explicit&0.025&46&12.79578&-4.24904&-&Yes&-1.25634\\7A AG&Explicit&0.0125&94&5.54182&5.48637&+&Yes&-1.1502\\
\hline
\end{tabular}
\caption{Proven periodic orbits for inverted pendulum model associated to the 7A analytic gradient controller.}\label{7A-orbit-table}
\end{table}

\begin{table}
\centering
\begin{tabular}{|c|c|c|c|c|c|c|c|c|}
\hline
Controller & Numerical Method & $h$ & $m$ & $\theta_0$ & $\omega_0$ & Direction & Exact $h$? & Max Reward  \\
\hline
9A AG&Explicit&0.05&25&12.67597&-3.53049&-&No&-0.89649\\9A AG&Explicit&0.05&25&6.17355&3.53059&+&No&-0.89651\\9A AG&Explicit&0.05&25&6.07665&3.80133&+&No&-0.90763\\9A AG&Explicit&0.025&54&13.28836&-4.90881&-&Yes&-0.60669\\9A AG&Explicit&0.025&54&5.5612&4.90881&+&Yes&-0.60669\\9A AG&Explicit&0.0125&118&13.52028&-5.45945&-&Yes&-0.4168\\9A AG&Explicit&0.0125&119&13.59144&-5.69248&-&No&-0.41451\\9A AG&Explicit&0.0125&119&5.25811&5.69246&+&No&-0.41451\\9A AG&Semi-Implicit&0.05&38&17.85968&-2.02001&-&Yes&-0.18072\\9A AG&Semi-Implicit&0.05&37&1.40931&3.17237&+&Yes&-0.19446\\9A AG&Semi-Implicit&0.05&37&12.85921&-3.10134&-&Yes&-0.19664\\9A AG&Semi-Implicit&0.05&35&19.65676&-5.18122&-&Yes&-0.23117\\9A AG&Semi-Implicit&0.05&38&17.27346&-3.56275&-&Yes&-0.18074\\9A AG&Semi-Implicit&0.025&72&4.46175&7.50944&+&Yes&-0.20408\\9A AG&Semi-Implicit&0.025&70&13.7473&-6.27949&-&Yes&-0.2233\\9A AG&Semi-Implicit&0.025&71&5.39232&5.254&+&Yes&-0.21325\\9A AG&Semi-Implicit&0.025&73&15.13705&-8.0&-&No&-0.20677\\9A AG&Semi-Implicit&0.025&71&13.95968&-6.8838&-&Yes&-0.21335\\9A AG&Semi-Implicit&0.0125&141&12.94968&-3.28075&-&Yes&-0.21434\\9A AG&Semi-Implicit&0.0125&140&14.63749&-7.71198&-&Yes&-0.21949\\9A AG&Semi-Implicit&0.0125&140&14.2623&-7.31164&-&Yes&-0.2195\\9A AG&Semi-Implicit&0.0125&140&4.58725&7.31164&+&Yes&-0.2195\\
\hline
\end{tabular}
\caption{Proven periodic orbits for the inverted pendulum model associated to the 9A analytic gradient controller. }\label{9A-orbit-table}
\end{table}

\begin{table}
\centering
\begin{tabular}{|c|c|c|c|c|c|c|c|c|}
\hline
Controller & Numerical Method & $h$ & $m$ & $\theta_0$ & $\omega_0$ & Direction & Exact $h$? & Max Reward  \\
\hline
13A AG&Explicit&0.05&26&13.50041&-5.82961&-&No&-0.79618\\13A AG&Explicit&0.05&26&5.9976&3.76616&+&No&-0.73857\\13A AG&Explicit&0.05&26&5.31309&5.92499&+&No&-0.78159\\13A AG&Explicit&0.05&26&13.84449&-6.75089&-&No&-0.79851\\13A AG&Explicit&0.05&26&5.05969&6.62089&+&No&-0.79601\\13A AG&Explicit&0.025&59&18.37965&-1.80891&-&No&-0.43973\\13A AG&Explicit&0.025&58&12.66676&-2.72797&-&Yes&-0.44836\\13A AG&Explicit&0.025&59&14.36394&-7.48261&-&No&-0.44529\\13A AG&Explicit&0.025&59&4.48575&7.48243&+&No&-0.44529\\13A AG&Explicit&0.0125&133&4.89743&6.59457&+&No&-0.26466\\13A AG&Explicit&0.0125&133&5.80166&3.64944&+&No&-0.26362\\13A AG&Explicit&0.0125&133&13.0479&-3.64944&-&No&-0.26362\\13A AG&Explicit&0.0125&133&5.80235&3.64672&+&No&-0.26343\\
\hline
\end{tabular}
\caption{Proven periodic orbits for the inverted pendulum model associated to the 13A analytic gradient controller. }
\end{table}

\begin{table}
\centering
\begin{tabular}{|c|c|c|c|c|c|c|c|c|}
\hline
Controller & Numerical Method & $h$ & $m$ & $\theta_0$ & $\omega_0$ & Direction & Exact $h$? & Max Reward  \\
\hline
17A AG&Explicit&0.05&26&1.90643&5.60505&+&Yes&-0.72076\\17A AG&Explicit&0.05&27&17.57395&-3.54371&-&No&-0.61191\\17A AG&Explicit&0.05&27&12.55099&-2.76102&-&No&-0.60389\\17A AG&Explicit&0.05&25&18.38712&-2.64432&-&Yes&-0.84185\\17A AG&Explicit&0.025&58&13.24405&-4.58198&-&Yes&-0.4476\\17A AG&Explicit&0.025&59&14.35807&-7.48942&-&No&-0.44369\\17A AG&Explicit&0.025&59&4.49148&7.48942&+&No&-0.44369\\17A AG&Explicit&0.025&59&4.67872&7.27377&+&No&-0.44352\\17A AG&Explicit&0.0125&135&5.27178&5.48681&+&Yes&-0.24559\\17A AG&Explicit&0.0125&135&5.28858&5.4297&+&Yes&-0.24547\\17A AG&Explicit&0.0125&135&13.4931&-5.19747&-&Yes&-0.24547\\
\hline
\end{tabular}
\caption{Proven periodic orbits for the inverted pendulum model associated to the 17A analytic gradient controller. }
\end{table}

\begin{table}
\centering
\begin{tabular}{|c|c|c|c|c|c|c|c|c|}
\hline
Controller & Numerical Method & $h$ & $m$ & $\theta_0$ & $\omega_0$ & Direction & Exact $h$? & Max Reward  \\
\hline
19A AG&Explicit&0.05&27&4.36177&7.70058&+&No&-0.67192\\19A AG&Explicit&0.05&27&13.42316&-5.45121&-&No&-0.66185\\19A AG&Explicit&0.05&26&5.70108&4.62699&+&Yes&-0.69967\\19A AG&Explicit&0.025&63&0.15916&1.76414&+&No&-0.31985\\19A AG&Explicit&0.025&63&12.92424&-3.32463&-&No&-0.32448\\19A AG&Explicit&0.025&63&12.76299&-2.78306&-&No&-0.32339\\19A AG&Explicit&0.025&63&12.77097&-2.81438&-&No&-0.3266\\19A AG&Explicit&0.025&63&6.1642&2.52864&+&No&-0.32021\\19A AG&Explicit&0.0125&173&12.77357&-2.41394&-&Yes&-0.12798\\19A AG&Explicit&0.0125&174&6.0492&2.51255&+&Yes&-0.12686\\19A AG&Explicit&0.0125&174&12.74993&-2.32242&-&Yes&-0.12684\\
\hline
\end{tabular}
\caption{Proven periodic orbits for the inverted pendulum model associated to the 19A analytic gradient controller. }
\end{table}

\begin{table}
\centering
\begin{tabular}{|c|c|c|c|c|c|c|c|c|}
\hline
Controller & Numerical Method & $h$ & $m$ & $\theta_0$ & $\omega_0$ & Direction & Exact $h$? & Max Reward  \\
\hline
7A CMA&Explicit&0.05&22&5.31337&6.57824&+&Yes&-1.49701\\7A CMA&Explicit&0.05&22&13.53618&-6.57824&-&Yes&-1.49701\\7A CMA&Explicit&0.025&47&4.42014&7.56331&+&Yes&-1.14512\\7A CMA&Explicit&0.025&47&14.24033&-7.35421&-&Yes&-1.14512\\7A CMA&Explicit&0.025&47&14.34022&-7.47212&-&Yes&-1.14564\\7A CMA&Explicit&0.0125&97&4.65398&7.23906&+&Yes&-1.00524\\
\hline
\end{tabular}
\caption{Proven periodic orbits for the inverted pendulum model associated to the 7A CMA controller. }
\end{table}

\begin{table}
\centering
\begin{tabular}{|c|c|c|c|c|c|c|c|c|}
\hline
Controller & Numerical Method & $h$ & $m$ & $\theta_0$ & $\omega_0$ & Direction & Exact $h$? & Max Reward  \\
\hline
9A CMA&Explicit&0.05&27&12.64045&-3.10454&-&No&-0.68384\\9A CMA&Explicit&0.05&26&0.59372&2.43828&+&Yes&-0.71503\\9A CMA&Explicit&0.05&26&18.49751&-2.43738&-&Yes&-0.71503\\9A CMA&Explicit&0.05&27&18.49709&-2.40759&-&No&-0.69822\\9A CMA&Explicit&0.05&27&0.0843&2.74474&+&No&-0.68356\\9A CMA&Explicit&0.025&60&18.80527&-2.22511&-&No&-0.39398\\9A CMA&Explicit&0.025&60&18.81127&-2.24969&-&No&-0.3983\\9A CMA&Explicit&0.025&60&0.07487&2.15632&+&No&-0.39751\\9A CMA&Explicit&0.025&60&18.75031&-2.09399&-&No&-0.39429\\9A CMA&Explicit&0.025&60&6.30641&2.28974&+&No&-0.39993\\9A CMA&Explicit&0.0125&141&13.38799&-4.79421&-&Yes&-0.21102\\9A CMA&Explicit&0.0125&141&5.04124&6.21195&+&Yes&-0.21102\\9A CMA&Explicit&0.0125&142&14.26742&-7.30306&-&Yes&-0.20612\\
\hline
\end{tabular}
\caption{Proven periodic orbits for the inverted pendulum model associated to the 9A CMA controller. }
\end{table}

\begin{table}
\centering
\begin{tabular}{|c|c|c|c|c|c|c|c|c|}
\hline
Controller & Numerical Method & $h$ & $m$ & $\theta_0$ & $\omega_0$ & Direction & Exact $h$? & Max Reward  \\
\hline
13A CMA&Explicit&0.05&27&5.08433&6.15206&+&No&-0.69597\\13A CMA&Explicit&0.05&27&13.76523&-6.15207&-&No&-0.69597\\13A CMA&Explicit&0.05&26&13.45594&-5.42196&-&Yes&-0.73061\\13A CMA&Explicit&0.025&59&13.28742&-4.59389&-&Yes&-0.43059\\13A CMA&Explicit&0.025&60&5.98973&3.25799&+&No&-0.41479\\13A CMA&Explicit&0.0125&133&4.75204&6.53762&+&Yes&-0.26716\\13A CMA&Explicit&0.0125&133&13.21806&-4.1709&-&Yes&-0.26716\\
\hline
\end{tabular}
\caption{Proven periodic orbits for the inverted pendulum model associated to the 13A CMA controller. }
\end{table}

\begin{table}
\centering
\begin{tabular}{|c|c|c|c|c|c|c|c|c|}
\hline
Controller & Numerical Method & $h$ & $m$ & $\theta_0$ & $\omega_0$ & Direction & Exact $h$? & Max Reward  \\
\hline
17A CMA&Explicit&0.05&29&4.06745&7.8094&+&Yes&-0.46523\\17A CMA&Explicit&0.05&28&0.26448&2.14379&+&Yes&-0.53353\\17A CMA&Explicit&0.05&27&5.24959&5.91354&+&Yes&-0.62505\\17A CMA&Explicit&0.05&27&13.78766&-6.47764&-&Yes&-0.61496\\17A CMA&Explicit&0.025&68&12.66814&-2.30752&-&Yes&-0.24383\\17A CMA&Explicit&0.025&69&14.87737&-7.80932&-&Yes&-0.23134\\17A CMA&Explicit&0.025&68&0.25548&1.32123&+&Yes&-0.24383\\17A CMA&Explicit&0.025&71&6.07584&2.6637&+&No&-0.2437\\
\hline
\end{tabular}
\caption{Proven periodic orbits for the inverted pendulum model associated to the 17A CMA controller.}
\end{table}

\begin{table}
\centering
\begin{tabular}{|c|c|c|c|c|c|c|c|c|}
\hline
Controller & Numerical Method & $h$ & $m$ & $\theta_0$ & $\omega_0$ & Direction & Exact $h$? & Max Reward  \\
\hline
19A CMA&Explicit&0.05&26&5.94086&3.95104&+&No&-0.74575\\19A CMA&Explicit&0.05&26&13.73332&-6.41827&-&No&-0.74697\\19A CMA&Explicit&0.05&26&6.14106&3.38929&+&No&-0.74394\\19A CMA&Explicit&0.05&26&6.13885&3.39668&+&No&-0.74478\\19A CMA&Explicit&0.05&26&12.7101&-3.39464&-&No&-0.74454\\19A CMA&Explicit&0.025&60&4.65803&7.16679&+&No&-0.40015\\19A CMA&Explicit&0.025&60&13.04806&-3.86307&-&No&-0.40257\\19A CMA&Explicit&0.025&60&12.70841&-2.76397&-&No&-0.40315\\19A CMA&Explicit&0.025&60&12.63885&-2.56135&-&No&-0.40402\\19A CMA&Explicit&0.0125&143&12.92031&-3.10373&-&No&-0.2065\\19A CMA&Explicit&0.0125&143&0.5733&1.16497&+&No&-0.20904\\19A CMA&Explicit&0.0125&143&0.1567&1.40983&+&No&-0.20843\\19A CMA&Explicit&0.0125&143&18.18449&-1.35702&-&No&-0.20898\\
\hline
\end{tabular}
\caption{Proven periodic orbits for the inverted pendulum model associated to the 19A CMA controller.}\label{19A-CMA-pendulum-orbits}
\end{table}

\begin{table}
\centering
\begin{tabular}{|c|c|c|c|c|c|c|}
\hline
Controller & Numerical Method & h & $\theta_0$ & $\omega_0$ & $T_p$ & Return\\
\hline 
7A AG&Semi-Implicit&0.05&2.72973&0.1981&50.0&-7483.29278\\7A AG&Semi-Implicit&0.05&-0.21485&4.45759&16.25&-1065.64015\\7A AG&Semi-Implicit&0.05&2.04462&-7.95743&15.8&-972.38496\\7A AG&Semi-Implicit&0.05&-39.45286&7.98095&15.65&-966.01785\\7A AG&Semi-Implicit&0.05&-1.69766&7.99871&15.65&-965.74433\\7A AG&Semi-Implicit&0.025&2.37066&-7.99083&18.5&-1966.62157\\7A AG&Semi-Implicit&0.025&-2.42555&7.99863&18.425&-1968.06541\\7A AG&Semi-Implicit&0.025&-16.42134&-7.99559&18.25&-1967.66079\\7A AG&Semi-Implicit&0.025&-2.42797&7.9998&18.35&-1966.6394\\7A AG&Semi-Implicit&0.025&8.71523&-7.99823&18.125&-1966.1161\\7A AG&Semi-Implicit&0.0125&3.84864&8.0&12.5&-4417.35966\\7A AG&Semi-Implicit&0.0125&2.43454&-8.0&12.5&-4417.35975\\7A AG&Semi-Implicit&0.0125&-2.43454&8.0&12.5&-4417.35975\\7A AG&Semi-Implicit&0.0125&-2.43454&8.0&12.5&-4417.35975\\7A AG&Semi-Implicit&0.0125&-2.43454&8.0&12.5&-4417.35976\\
\hline
\end{tabular}
\caption{Proven persistent solutions for the inverted pendulum model associated to the 7A analytic gradient controller.}\label{7A-persistent-pendulum}
\end{table}

\begin{table}
\centering
\begin{tabular}{|c|c|c|c|c|c|c|}
\hline
Controller & Numerical Method & h & $\theta_0$ & $\omega_0$ & $T_p$ & Return\\
\hline 
13A AG&Semi-Implicit&0.05&3.1412&0.00377&8.1&-459.94732\\13A AG&Semi-Implicit&0.05&-3.12597&0.09801&10.7&-525.12367\\13A AG&Semi-Implicit&0.05&3.12008&0.30463&9.35&-454.33445\\13A AG&Semi-Implicit&0.05&-3.20741&-0.97794&7.8&-304.06445\\13A AG&Semi-Implicit&0.05&-46.94919&1.29085&5.5&-403.15886\\13A AG&Semi-Implicit&0.025&3.14159&0.0&11.4&-1811.38974\\13A AG&Semi-Implicit&0.025&3.14762&0.05012&8.5&-825.08973\\13A AG&Semi-Implicit&0.025&-3.15353&-0.08404&10.2&-1053.09524\\13A AG&Semi-Implicit&0.025&3.12534&-0.10417&10.0&-1034.64048\\13A AG&Semi-Implicit&0.025&-3.17807&-0.13783&10.025&-976.88593\\13A AG&Semi-Implicit&0.0125&-3.14158&6.0e-5&10.775&-2848.25762\\13A AG&Semi-Implicit&0.0125&3.13961&-0.01999&10.35&-2259.38486\\13A AG&Semi-Implicit&0.0125&3.14479&0.02947&10.0125&-2218.09884\\13A AG&Semi-Implicit&0.0125&3.18661&0.1071&8.4125&-1390.97149\\13A AG&Semi-Implicit&0.0125&-3.18753&0.0672&10.0125&-1862.06843\\
\hline
\end{tabular}
\caption{Proven persistent solutions for the inverted pendulum model associated to the 13A analytic gradient controller. }
\end{table}

\begin{table}
\centering
\begin{tabular}{|c|c|c|c|c|c|c|}
\hline
Controller & Numerical Method & h & $\theta_0$ & $\omega_0$ & $T_p$ & Return\\
\hline 
17A AG&Semi-Implicit&0.05&-21.98237&0.01005&11.75&-870.28774\\17A AG&Semi-Implicit&0.05&-3.03105&0.98335&9.8&-678.21786\\17A AG&Semi-Implicit&0.05&-5.98521&7.29977&12.5&-798.03779\\17A AG&Semi-Implicit&0.05&-3.5797&0.74532&11.65&-742.87019\\17A AG&Semi-Implicit&0.05&3.14602&-0.17762&9.4&-711.1134\\17A AG&Semi-Implicit&0.025&3.14159&0.0&8.675&-2085.36882\\17A AG&Semi-Implicit&0.025&3.14159&0.0&10.075&-1793.36929\\17A AG&Semi-Implicit&0.025&3.14159&-0.0&10.525&-1642.56476\\17A AG&Semi-Implicit&0.025&-3.14143&0.00114&10.125&-1242.74263\\17A AG&Semi-Implicit&0.025&-3.12133&0.48934&7.9&-632.39914\\17A AG&Semi-Implicit&0.0125&3.14158&-7.0e-5&10.25&-2817.79332\\17A AG&Semi-Implicit&0.0125&-21.99116&-9.0e-5&10.1875&-2785.64168\\17A AG&Semi-Implicit&0.0125&3.14281&0.00755&10.5625&-2272.13871\\17A AG&Semi-Implicit&0.0125&-3.13898&0.03902&5.4375&-1857.35504\\17A AG&Semi-Implicit&0.0125&3.13852&-0.33796&9.55&-1804.3914\\
\hline
\end{tabular}
\caption{Proven persistent solutions for the inverted pendulum model associated to the 17A analytic gradient controller.}
\end{table}

\begin{table}
\centering
\begin{tabular}{|c|c|c|c|c|c|c|}
\hline
Controller & Numerical Method & h & $\theta_0$ & $\omega_0$ & $T_p$ & Return\\
\hline 
19A AG&Semi-Implicit&0.05&-3.14159&-0.0&13.9&-2010.05434\\19A AG&Semi-Implicit&0.05&3.14159&0.0&12.65&-1992.49113\\19A AG&Semi-Implicit&0.05&-3.14159&-0.0&14.15&-1562.49732\\19A AG&Semi-Implicit&0.05&3.14212&0.00766&10.6&-779.12907\\19A AG&Semi-Implicit&0.05&3.13976&-0.01001&10.0&-625.92422\\19A AG&Semi-Implicit&0.025&-3.14159&-0.0&13.475&-2600.22201\\19A AG&Semi-Implicit&0.025&-3.1416&-4.0e-5&12.15&-2184.2091\\19A AG&Semi-Implicit&0.025&-3.14694&0.11007&9.025&-864.14419\\19A AG&Semi-Implicit&0.025&-3.08596&1.10157&9.4&-855.29\\19A AG&Semi-Implicit&0.025&-3.29041&2.44094&8.75&-417.86503\\19A AG&Semi-Implicit&0.0125&-3.1419&-0.00079&11.4125&-3338.67144\\19A AG&Semi-Implicit&0.0125&-3.13857&0.03648&10.375&-2581.50326\\19A AG&Semi-Implicit&0.0125&3.13463&-0.04383&10.45&-2509.19096\\19A AG&Semi-Implicit&0.0125&3.12904&-0.20442&10.25&-2147.51587\\19A AG&Semi-Implicit&0.0125&3.08347&-0.26131&9.725&-2004.57505\\
\hline
\end{tabular}
\caption{Proven persistent solutions for the inverted pendulum model associated to the 19A analytic gradient controller. }
\end{table}

\begin{table}
\centering
\begin{tabular}{|c|c|c|c|c|c|c|}
\hline
Controller & Numerical Method & h & $\theta_0$ & $\omega_0$ & $T_p$ & Return\\
\hline 
% 7A CMA&Semi-Implicit&0.05&0.85628&-7.83273&14.9&-1083.01397\\7A CMA&Semi-Implicit&0.05&0.29723&4.21378&15.6&-930.23824\\7A CMA&Semi-Implicit&0.05&2.71757&0.19875&50.0&-7481.07111\\7A CMA&Semi-Implicit&0.05&-1.18097&7.90936&15.5&-1084.74031\\7A CMA&Semi-Implicit&0.05&0.28434&-7.72116&15.6&-948.4485\\7A CMA&Semi-Implicit&0.025&2.67909&0.09195&25.0&-7500.96493\\7A CMA&Semi-Implicit&0.025&-2.68938&-0.14449&25.0&-7504.31682\\7A CMA&Semi-Implicit&0.025&-0.59568&5.58435&13.3&-1607.13538\\7A CMA&Semi-Implicit&0.025&2.13619&-7.91718&13.5&-1666.45407\\7A CMA&Semi-Implicit&0.025&2.68287&0.11514&25.0&-7493.30914$\dagger$\\7A CMA&Semi-Implicit&0.0125&2.38441&-7.99882&12.5&-3361.25771*\\7A CMA&Semi-Implicit&0.0125&-3.88133&-7.99999&12.5&-3355.99829\\7A CMA&Semi-Implicit&0.0125&-8.66725&7.9985&12.5&-3358.36131\\7A CMA&Semi-Implicit&0.0125&-2.37438&7.99604&12.5&-3367.29765\\7A CMA&Semi-Implicit&0.0125&3.52343&-0.18886&12.5&-7526.6398\\
7A CMA&Explicit&0.0125&-40.41906&-0.12086&12.5&-6319.37599\\7A CMA&Semi-Implicit&0.05&-2.73051&-0.20834&50.0&-7483.55621\\7A CMA&Semi-Implicit&0.05&-3.55455&0.20764&50.0&-7483.43715\\7A CMA&Semi-Implicit&0.05&-2.72436&-0.20515&50.0&-7483.02034\\7A CMA&Semi-Implicit&0.05&2.71877&0.20012&50.0&-7482.63408\\7A CMA&Semi-Implicit&0.05&1.01554&-7.98294&16.9&-957.89294\\7A CMA&Semi-Implicit&0.025&-2.69323&-0.158&25.0&-7505.32605\\7A CMA&Semi-Implicit&0.025&2.24532&-7.99685&15.325&-1684.98931\\7A CMA&Semi-Implicit&0.025&-2.24572&7.99676&14.925&-1684.05543\\7A CMA&Semi-Implicit&0.025&-8.52876&7.99679&14.675&-1681.82397\\7A CMA&Semi-Implicit&0.025&2.26987&-7.99202&13.225&-1671.0005\\\color{blue}7A CMA&\color{blue}Semi-Implicit&\color{blue}0.0125&\color{blue}-3.50688&\color{blue}0.13596&\color{blue}12.5&\color{blue}-7527.4697\\7A CMA&Semi-Implicit&0.0125&-2.37537&7.99694&12.5&-3378.62298\\7A CMA&Semi-Implicit&0.0125&-2.38556&7.9997&12.5&-2859.47861\\7A CMA&Semi-Implicit&0.0125&2.37753&-7.99888&12.5&-3368.21153\\7A CMA&Semi-Implicit&0.0125&2.38589&-7.99995&12.5&-3365.43752\\
\hline
\end{tabular}
\caption{Proven persistent solutions for the inverted pendulum model associated to the 7A CMA controller.
% The symbol * indicates the transient is plotted in Figure \ref{fig:transient1}. 
The row with blue text references the solution plotted in Figure \ref{fig:transient2}.} \label{7A_CMA_transients}
\end{table}

\begin{table}
\centering
\begin{tabular}{|c|c|c|c|c|c|c|}
\hline
Controller & Numerical Method & h & $\theta_0$ & $\omega_0$ & $T_p$ & Return\\
\hline 
9A CMA&Semi-Implicit&0.05&-3.14159&-0.0&9.75&-1338.82442\\9A CMA&Semi-Implicit&0.05&3.14159&0.0&10.85&-1409.0304\\9A CMA&Semi-Implicit&0.05&-3.14159&-3.0e-5&9.3&-1087.74734\\9A CMA&Semi-Implicit&0.05&3.14155&-0.0&8.85&-910.06371\\9A CMA&Semi-Implicit&0.05&3.14146&-0.00078&8.95&-836.8964\\9A CMA&Semi-Implicit&0.025&3.14159&0.0&11.525&-2812.38093\\9A CMA&Semi-Implicit&0.025&-3.14159&0.0&9.575&-2516.36021\\9A CMA&Semi-Implicit&0.025&3.14159&-0.0&9.525&-2512.81421\\9A CMA&Semi-Implicit&0.025&3.14159&-2.0e-5&10.575&-2399.72389\\9A CMA&Semi-Implicit&0.025&-3.14159&1.0e-5&10.95&-2311.50848\\ \color{blue}9A CMA& \color{blue}Semi-Implicit& \color{blue}0.0125& \color{blue}3.14159& \color{blue}0.0& \color{blue}9.8875& \color{blue}-5031.73384\\9A CMA&Semi-Implicit&0.0125&3.14159&0.0&9.2375&-4965.00622\\9A CMA&Semi-Implicit&0.0125&-3.14159&-0.0&9.75&-5092.36065\\9A CMA&Semi-Implicit&0.0125&3.14159&0.0&8.8375&-4893.50163\\9A CMA&Semi-Implicit&0.0125&-3.14159&0.0&8.4125&-4751.56075\\
\hline
\end{tabular}
\caption{Proven persistent solutions for the inverted pendulum model associated to the 9A CMA controller. The row with blue text references the solution plotted in Figure \ref{fig:transient2_wrapping}. }\label{9A_CMA_transients}
\end{table}

\begin{table}
\centering
\begin{tabular}{|c|c|c|c|c|c|c|}
\hline
Controller & Numerical Method & h & $\theta_0$ & $\omega_0$ & $T_p$ & Return\\
\hline 
13A CMA&Semi-Implicit&0.05&-3.14159&-0.0&8.45&-1148.62728\\13A CMA&Semi-Implicit&0.05&3.14159&0.0&9.6&-1215.87481\\13A CMA&Semi-Implicit&0.05&-3.14159&0.0&9.2&-1201.32983\\13A CMA&Semi-Implicit&0.05&-3.14159&-0.0&8.4&-1111.26135\\13A CMA&Semi-Implicit&0.05&-3.14154&6.0e-5&9.9&-931.12334\\13A CMA&Semi-Implicit&0.025&3.14159&0.0&9.275&-2332.03277\\13A CMA&Semi-Implicit&0.025&3.14159&-0.0&8.05&-2134.44535\\13A CMA&Semi-Implicit&0.025&3.14162&0.00054&9.575&-1742.59427\\13A CMA&Semi-Implicit&0.025&-3.14425&-0.04766&9.575&-1290.86516\\13A CMA&Semi-Implicit&0.025&3.16163&0.07569&10.125&-1203.76984\\13A CMA&Semi-Implicit&0.0125&-3.14159&-1.0e-5&9.8875&-4268.54038\\13A CMA&Semi-Implicit&0.0125&-3.1423&-0.00649&9.8875&-2958.44871\\13A CMA&Semi-Implicit&0.0125&3.119&-0.03818&9.9375&-2341.27276\\13A CMA&Semi-Implicit&0.0125&-3.11048&0.49471&10.2875&-2128.33919\\13A CMA&Semi-Implicit&0.0125&3.06352&-0.6774&8.1&-1546.66125\\
\hline
\end{tabular}
\caption{Proven persistent solutions for the inverted pendulum model associated to the 13A CMA controller. }
\end{table}

\begin{table}
\centering
\begin{tabular}{|c|c|c|c|c|c|c|}
\hline
Controller & Numerical Method & h & $\theta_0$ & $\omega_0$ & $T_p$ & Return\\
\hline 
17A CMA&Explicit&0.0125&-3.14279&-0.00463&6.0875&-2113.98781\\17A CMA&Explicit&0.0125&3.14&-0.00619&6.175&-2175.54243\\17A CMA&Explicit&0.0125&3.13994&-0.00644&6.2&-2191.09273\\17A CMA&Explicit&0.0125&-3.13983&0.00684&7.4&-2528.584\\17A CMA&Explicit&0.0125&-3.14341&-0.00708&6.2&-2187.88613\\17A CMA&Semi-Implicit&0.05&-3.14028&0.0057&7.8&-673.88216\\17A CMA&Semi-Implicit&0.05&3.1387&-0.01248&8.3&-637.66252\\17A CMA&Semi-Implicit&0.05&3.14646&0.02088&8.9&-623.71155\\17A CMA&Semi-Implicit&0.05&-3.15392&-0.05135&9.6&-587.25306\\17A CMA&Semi-Implicit&0.05&-3.13291&0.03643&8.2&-591.69433\\17A CMA&Semi-Implicit&0.025&3.14213&0.00218&7.925&-1400.80396\\17A CMA&Semi-Implicit&0.025&-3.14103&0.0023&7.125&-1290.77127\\17A CMA&Semi-Implicit&0.025&-3.1434&-0.00734&8.275&-1072.70888\\17A CMA&Semi-Implicit&0.025&3.14494&0.01356&8.55&-1264.84976\\17A CMA&Semi-Implicit&0.025&-3.1373&0.01739&7.95&-991.04271\\17A CMA&Semi-Implicit&0.0125&-3.14092&0.00266&7.15&-2612.43124\\17A CMA&Semi-Implicit&0.0125&3.1394&-0.00869&7.8375&-2576.58937\\17A CMA&Semi-Implicit&0.0125&-3.1386&0.01185&7.975&-2047.44841\\17A CMA&Semi-Implicit&0.0125&3.14709&0.02175&8.925&-2437.95276\\17A CMA&Semi-Implicit&0.0125&-3.13499&0.02611&8.6&-2398.39817\\
\hline
\end{tabular}
\caption{Proven persistent solutions for the inverted pendulum model associated to the 17A CMA controller. }
\end{table}

\begin{table}
\centering
\begin{tabular}{|c|c|c|c|c|c|c|}
\hline
Controller & Numerical Method & h & $\theta_0$ & $\omega_0$ & $T_p$ & Return\\
\hline 
19A CMA&Explicit&0.0125&-78.75089&0.3764&12.5&-4044.82608\\19A CMA&Semi-Implicit&0.05&3.15467&0.03093&7.65&-597.79139\\19A CMA&Semi-Implicit&0.05&3.15723&0.03698&7.3&-585.09022\\19A CMA&Semi-Implicit&0.05&-3.16343&-0.05166&7.0&-581.37866\\19A CMA&Semi-Implicit&0.05&3.11608&-0.06037&6.55&-510.8697\\19A CMA&Semi-Implicit&0.05&3.1051&-0.08642&5.5&-548.00379\\19A CMA&Semi-Implicit&0.025&-3.16664&-0.05896&7.1&-1118.10272\\19A CMA&Semi-Implicit&0.025&3.10346&-0.08981&6.65&-1083.10753\\19A CMA&Semi-Implicit&0.025&-3.18558&-0.10366&6.4&-1065.56841\\19A CMA&Semi-Implicit&0.025&-3.08897&0.12409&5.7&-1040.56357\\19A CMA&Semi-Implicit&0.025&3.19591&0.12812&6.225&-1034.15971\\19A CMA&Semi-Implicit&0.0125&-3.11073&0.07248&7.125&-2146.97483\\19A CMA&Semi-Implicit&0.0125&3.17691&0.08297&6.2375&-1858.22941\\19A CMA&Semi-Implicit&0.0125&3.18517&0.10243&6.175&-1806.74824\\19A CMA&Semi-Implicit&0.0125&3.19068&0.11544&6.6375&-2081.32522\\19A CMA&Semi-Implicit&0.0125&-3.1908&-0.11572&6.3875&-2092.97931\\
\hline
\end{tabular}
\caption{Proven persistent solutions for the inverted pendulum model associated to the 19A CMA controller. }\label{19A-persistent-pendulum}
\end{table}

\section{Proven periodic orbits and persistent solutions for cart-pole swingup}\label{sec:proven_orbits_cartpole}
In the following pages, we catalogue the periodic orbits and persistent solutions we have proven for the cart-pole swingup problem. To improve readability, all initial angles, angular velocities and step sizes are truncated to three decimal places. In all cases, numbered controllers reference those in Table \ref{table_controllers_cartpole}. The periodic orbits (Table \ref{cartpoleswingup-deepnet}) were all proven for un-fixed step size. We include the period $m$, step size $h$, mean penalty values, maximum amplitude $|\theta|$, and $\epsilon$ level at which the period orbit classifies as persistent. For persistent solutions (Table \ref{cartpoleswing-persistent-smallnet} through Table \ref{cartpoleswing-persistent-21}), we integrate forward from ICs $(x_0,\dot x_0,\theta_0,\dot\theta_0)$ with the specified integrator. Note that the initial conditions have been rounded for readability, hence the appearance of duplicates in the tables. The field ``Escaped?" refers to whether or not the cart escapes the domain $|x|\leq 2.4$ within 2000 episode steps. All numerical quantities (aside from $m$) are rounded midpoints of intervals that contain the true values. The radii of these intervals are in all cases small, typically around $10^{-14}$.

\begin{table}
\centering
\begin{tabular}{|c|l|}
\hline
Controller & Formula\\
\hline
symb.\ $k=17$ & $ ((x_3 \cdot 92.07) + 35.31\cdot x_4) / (((x_4 \cdot ((x_3 \cdot 14.61) + 2.56 \cdot x_4)) \cdot  -3.52)  -12.62) $\\
symb.\ $k=19$ & $ ((x_3 \cdot 5.04) + 1.42 \cdot x_4) / ((((-1.83 \cdot x_4 + 1.35 \cdot x_3) \cdot ((x_3 \cdot 3.35) + 0.50\cdot x_4)) \cdot 0.33)  -1.15)$\\
symb.\ $k=21$ & $ ((x_3 \cdot 6.76) + 3.62 \cdot x_4) / (((((x_3 \cdot 3.25) + 0.66 \cdot x_4) \cdot ((x_3 \cdot 9.13) + 1.20 \cdot x_4)) \cdot -0.75) + -0.14) $
\\
\hline
\end{tabular}
\caption{Symbolic controllers dictionary for cart-pole swingup. For readability, all parameters are truncated to two decimal places. Note that $x_3=\sin(\theta)$ and $x_4=\dot\theta$. }\label{table_controllers_cartpole}
\end{table}

\begin{table}
\centering
\begin{tabular}{|c|c|c|c|c|c|c|}
\hline
Controller & Num.\ Method & $h$ & $m$ & Mean Penalty & Maximum $|\theta|$ & Persistent $\epsilon$ level\\
\hline
Deep NN & Explicit & 0.009808 & 74 & -0.8918 & 0.06137 & 0.04215 \\
Deep NN & Semi-Implicit & 0.009821 & 69 & -0.9136 & 0.05198 & 0.03606\\
Deep NN & Explicit & 0.0004972 & 141 & -0.8985 & 0.05751 & 0.04048\\
Deep NN & Semi-Implicit & 0.0004909 & 138 & -0.9098 & 0.05282 & 0.03723 \\
\hline
\end{tabular}
\caption{Proven periodic orbits for the cart-pole swingup model associated to the Deep NN model.}\label{cartpoleswingup-deepnet}
\end{table}

\begin{table}
\centering
\begin{tabular}{|c|c|c|c|c|c|c|c|c|c}
\hline
Controller & Num.\ Method & $h$ & $x_0$ & $\dot x_0$ & $\theta_0$ & $\dot\theta_0$ & Escaped? & Acc.\ Pen.\ \\
\hline
Small NN&Explicit&0.01&0.488&0.5&2.642&0.498&Yes&519.117\\Small NN&Explicit&0.01&0.5&0.496&2.642&0.5&Yes&514.321\\Small NN&Explicit&0.01&0.492&0.5&2.642&0.5&Yes&496.491\\Small NN&Explicit&0.01&0.442&0.5&2.642&0.5&Yes&510.154\\Small NN&Explicit&0.01&0.491&0.499&2.647&0.499&Yes&504.107\\Small NN&Explicit&0.005&0.422&0.5&2.642&0.5&Yes&2343.409\\Small NN&Explicit&0.005&0.423&0.5&2.642&0.5&Yes&2343.111\\Small NN&Explicit&0.005&0.422&0.5&2.642&0.5&Yes&2342.457\\Small NN&Explicit&0.005&0.318&0.5&2.642&0.5&Yes&2341.617\\Small NN&Explicit&0.005&0.319&0.5&2.642&0.5&Yes&2343.503\\Small NN&Explicit&0.0025&0.301&0.5&2.642&0.5&Yes&6263.193\\Small NN&Explicit&0.0025&0.301&0.5&2.642&0.5&Yes&6263.193\\Small NN&Explicit&0.0025&0.301&0.5&2.642&0.5&Yes&6263.193\\Small NN&Explicit&0.0025&0.301&0.5&2.642&0.5&Yes&6263.194\\Small NN&Explicit&0.0025&0.301&0.5&2.642&0.5&Yes&6263.192\\Small NN&Semi-Implicit&0.01&-0.36&-0.498&3.116&-0.483&Yes&2.684$\times 10^6$\\Small NN&Semi-Implicit&0.01&0.5&0.5&2.642&0.498&Yes&-6.224\\Small NN&Semi-Implicit&0.01&0.394&0.5&2.642&0.499&Yes&0.711\\Small NN&Semi-Implicit&0.01&0.496&0.5&2.642&0.5&Yes&-0.23\\Small NN&Semi-Implicit&0.01&0.499&0.5&2.642&0.498&Yes&-0.215\\Small NN&Semi-Implicit&0.005&-0.463&-0.485&3.277&-0.499&Yes&7.984$\times 10^6$\\Small NN&Semi-Implicit&0.005&-0.498&-0.457&3.145&-0.482&Yes&7.71$\times 10^6$\\Small NN&Semi-Implicit&0.005&-0.188&-0.488&2.939&-0.141&Yes&43847.201\\Small NN&Semi-Implicit&0.005&0.357&0.5&2.642&0.5&Yes&1875.438\\Small NN&Semi-Implicit&0.005&0.44&0.5&2.642&0.5&Yes&1874.277\\Small NN&Semi-Implicit&0.0025&-0.442&-0.497&3.258&-0.491&Yes&3.63$\times 10^6$\\Small NN&Semi-Implicit&0.0025&-0.431&-0.5&3.228&-0.468&Yes&3.52$\times 10^6$\\Small NN&Semi-Implicit&0.0025&-0.497&-0.482&3.105&-0.471&Yes&3.448$\times 10^6$\\Small NN&Semi-Implicit&0.0025&0.001&-0.409&3.104&-0.245&Yes&23101.909\\Small NN&Semi-Implicit&0.0025&0.317&0.5&2.642&0.5&Yes&5800.631\\
\hline
\end{tabular}
\caption{Proven persistent solutions for the cart-pole swingup model associated to small NN controller. }\label{cartpoleswing-persistent-smallnet}
\end{table}

\begin{table}
\centering
\begin{tabular}{|c|c|c|c|c|c|c|c|c|c}
\hline
Controller & Num.\ Method & $h$ & $x_0$ & $\dot x_0$ & $\theta_0$ & $\dot\theta_0$ & Escaped? & Acc.\ Pen.\ \\
\hline
symb.\ $k=17$&Explicit&0.01&0.455&-0.5&3.633&-0.497&Yes&5398.721\\symb.\ $k=17$&Explicit&0.01&0.292&0.5&2.65&0.497&Yes&5398.721\\symb.\ $k=17$&Explicit&0.01&-0.21&0.5&2.65&0.497&Yes&4171.181\\symb.\ $k=17$&Explicit&0.01&-0.22&-0.5&3.633&-0.497&Yes&4171.293\\symb.\ $k=17$&Explicit&0.01&0.334&-0.5&3.633&-0.497&Yes&4170.587\\symb.\ $k=17$&Explicit&0.005&-0.481&-0.5&3.642&-0.479&Yes&5559.668\\symb.\ $k=17$&Explicit&0.005&0.498&0.5&2.642&0.5&Yes&5558.377\\symb.\ $k=17$&Explicit&0.005&0.151&0.5&2.642&0.5&Yes&5558.377\\symb.\ $k=17$&Explicit&0.005&-0.065&-0.5&3.642&-0.5&Yes&5558.377\\symb.\ $k=17$&Explicit&0.005&-0.212&0.5&2.642&0.5&Yes&5558.377\\symb.\ $k=17$&Explicit&0.0025&-0.208&-0.5&3.639&-0.499&Yes&9948.271\\symb.\ $k=17$&Explicit&0.0025&0.142&0.5&2.642&0.5&Yes&9945.919\\symb.\ $k=17$&Explicit&0.0025&0.113&-0.5&3.642&-0.5&Yes&9945.919\\symb.\ $k=17$&Explicit&0.0025&-0.316&-0.5&3.642&-0.5&Yes&9945.919\\symb.\ $k=17$&Explicit&0.0025&0.5&0.5&2.642&0.5&Yes&9945.919\\symb.\ $k=17$&Semi-Implicit&0.01&0.498&0.499&2.973&-0.499&Yes&19088.315\\symb.\ $k=17$&Semi-Implicit&0.01&0.455&0.499&2.977&-0.49&Yes&18389.379\\symb.\ $k=17$&Semi-Implicit&0.01&0.023&0.49&2.975&-0.495&Yes&18010.231\\symb.\ $k=17$&Semi-Implicit&0.01&-0.36&0.453&2.975&-0.495&Yes&17469.444\\symb.\ $k=17$&Semi-Implicit&0.01&0.498&-0.5&3.309&0.497&Yes&17739.411\\symb.\ $k=17$&Semi-Implicit&0.005&0.5&-0.497&3.307&0.492&Yes&14403.195\\symb.\ $k=17$&Semi-Implicit&0.005&-0.133&-0.485&3.31&0.5&Yes&13950.51\\symb.\ $k=17$&Semi-Implicit&0.005&0.486&0.484&2.977&-0.489&Yes&13899.735\\symb.\ $k=17$&Semi-Implicit&0.005&0.493&-0.499&3.308&0.495&Yes&12975.006\\symb.\ $k=17$&Semi-Implicit&0.005&-0.121&0.493&2.973&-0.499&Yes&11972.97\\symb.\ $k=17$&Semi-Implicit&0.0025&-0.225&0.414&2.987&-0.495&Yes&9847.417\\symb.\ $k=17$&Semi-Implicit&0.0025&-0.469&0.428&3.005&-0.475&Yes&9479.113\\symb.\ $k=17$&Semi-Implicit&0.0025&-0.244&-0.218&3.282&0.421&Yes&9284.573\\symb.\ $k=17$&Semi-Implicit&0.0025&-0.251&0.5&2.642&0.5&Yes&9231.592\\symb.\ $k=17$&Semi-Implicit&0.0025&0.163&-0.5&3.642&-0.5&Yes&9231.592\\
\hline
\end{tabular}
\caption{Proven persistent solutions for the cart-pole swingup model associated to the $k=17$ symbolic controller. }\label{cartpoleswing-persistent-17}
\end{table}

\begin{table}
\centering
\begin{tabular}{|c|c|c|c|c|c|c|c|c|c}
\hline
Controller & Num.\ Method & $h$ & $x_0$ & $\dot x_0$ & $\theta_0$ & $\dot\theta_0$ & Escaped? & Acc.\ Pen.\ \\
\hline
symb.\ $k=19$&Explicit&0.01&-0.416&0.5&2.642&0.5&Yes&8768.342\\symb.\ $k=19$&Explicit&0.01&-0.402&0.5&2.642&0.5&Yes&8768.342\\symb.\ $k=19$&Explicit&0.01&-0.371&-0.5&3.642&-0.5&Yes&8768.342\\symb.\ $k=19$&Explicit&0.01&-0.414&0.5&2.642&0.5&Yes&8768.342\\symb.\ $k=19$&Explicit&0.01&-0.121&0.5&2.642&0.5&Yes&8768.342\\symb.\ $k=19$&Explicit&0.005&0.483&-0.5&3.642&-0.5&Yes&7259.325\\symb.\ $k=19$&Explicit&0.005&0.186&-0.5&3.642&-0.5&Yes&7259.325\\symb.\ $k=19$&Explicit&0.005&0.091&0.5&2.642&0.5&Yes&7259.325\\symb.\ $k=19$&Explicit&0.005&-0.316&0.5&2.642&0.5&Yes&7259.325\\symb.\ $k=19$&Explicit&0.005&-0.318&-0.5&3.642&-0.5&Yes&7259.325\\symb.\ $k=19$&Explicit&0.0025&-0.499&-0.5&3.642&-0.499&Yes&11062.468\\symb.\ $k=19$&Explicit&0.0025&-0.014&-0.5&3.642&-0.499&Yes&11062.468\\symb.\ $k=19$&Explicit&0.0025&-0.193&0.5&2.642&0.499&Yes&11062.468\\symb.\ $k=19$&Explicit&0.0025&-0.144&-0.5&3.642&-0.499&Yes&11062.468\\symb.\ $k=19$&Explicit&0.0025&-0.474&0.5&2.642&0.499&Yes&11062.468\\symb.\ $k=19$&Semi-Implicit&0.01&0.065&-0.499&3.245&0.5&Yes&29016.936\\symb.\ $k=19$&Semi-Implicit&0.01&-0.374&0.493&3.044&-0.472&Yes&29532.913\\symb.\ $k=19$&Semi-Implicit&0.01&0.338&-0.495&3.235&0.455&Yes&29034.726\\symb.\ $k=19$&Semi-Implicit&0.01&-0.012&-0.473&3.241&0.483&Yes&28731.003\\symb.\ $k=19$&Semi-Implicit&0.01&0.4&-0.474&3.245&0.499&Yes&28477.151\\symb.\ $k=19$&Semi-Implicit&0.005&0.477&0.498&3.052&-0.433&Yes&20459.818\\symb.\ $k=19$&Semi-Implicit&0.005&-0.36&0.5&3.038&-0.499&Yes&19816.419\\symb.\ $k=19$&Semi-Implicit&0.005&0.315&0.497&3.054&-0.423&Yes&19060.325\\symb.\ $k=19$&Semi-Implicit&0.005&0.04&0.443&3.048&-0.452&Yes&15495.356\\symb.\ $k=19$&Semi-Implicit&0.005&-0.427&-0.489&3.237&0.484&Yes&8075.266\\symb.\ $k=19$&Semi-Implicit&0.0025&-0.007&-0.5&3.642&-0.484&Yes&10192.822\\symb.\ $k=19$&Semi-Implicit&0.0025&-0.345&0.5&2.642&0.484&Yes&10192.822\\symb.\ $k=19$&Semi-Implicit&0.0025&0.017&-0.5&3.642&-0.484&Yes&10192.822\\symb.\ $k=19$&Semi-Implicit&0.0025&0.385&0.5&2.642&0.484&Yes&10192.822\\symb.\ $k=19$&Semi-Implicit&0.0025&-0.483&0.5&2.642&0.484&Yes&10192.822\\
\hline
\end{tabular}
\caption{Proven persistent solutions for the cart-pole swingup model associated to the $k=19$ symbolic controller. }
\end{table}

\begin{table}
\centering
\begin{tabular}{|c|c|c|c|c|c|c|c|c|c}
\hline
Controller & Num.\ Method & $h$ & $x_0$ & $\dot x_0$ & $\theta_0$ & $\dot\theta_0$ & Escaped? & Acc.\ Pen.\ \\
\hline
symb.\ $k=21$&Explicit&0.01&-0.449&0.498&2.9&-0.498&Yes&26674.106\\symb.\ $k=21$&Explicit&0.01&-0.354&-0.441&3.364&0.458&Yes&23714.699\\symb.\ $k=21$&Explicit&0.01&0.474&0.499&2.906&-0.5&Yes&22918.827\\symb.\ $k=21$&Explicit&0.01&0.263&-0.5&3.373&0.493&Yes&22907.945\\symb.\ $k=21$&Explicit&0.01&-0.2&0.499&2.908&-0.498&Yes&25699.503\\symb.\ $k=21$&Explicit&0.005&0.497&-0.441&3.382&0.495&Yes&8276.237\\symb.\ $k=21$&Explicit&0.005&0.053&0.497&2.937&-0.422&Yes&7591.779\\symb.\ $k=21$&Explicit&0.005&0.5&-0.5&3.348&0.453&Yes&7002.795\\symb.\ $k=21$&Explicit&0.005&-0.124&-0.124&2.932&-0.425&Yes&5203.709\\symb.\ $k=21$&Explicit&0.005&0.243&0.5&3.513&0.5&Yes&5232.379\\symb.\ $k=21$&Explicit&0.0025&-0.192&0.5&2.642&0.475&Yes&11611.265\\symb.\ $k=21$&Explicit&0.0025&0.075&0.5&2.642&0.476&Yes&11611.505\\symb.\ $k=21$&Explicit&0.0025&-0.374&0.5&2.642&0.476&Yes&11611.548\\symb.\ $k=21$&Explicit&0.0025&0.457&0.5&2.642&0.476&Yes&11610.95\\symb.\ $k=21$&Explicit&0.0025&0.364&0.5&2.642&0.476&Yes&11611.447\\symb.\ $k=21$&Semi-Implicit&0.01&-0.014&-0.499&3.377&0.495&Yes&37746.083\\symb.\ $k=21$&Semi-Implicit&0.01&0.177&-0.491&3.374&0.49&Yes&37760.939\\symb.\ $k=21$&Semi-Implicit&0.01&0.378&-0.5&3.375&0.492&Yes&34717.066\\symb.\ $k=21$&Semi-Implicit&0.01&0.217&0.483&2.906&-0.487&Yes&37943.141\\symb.\ $k=21$&Semi-Implicit&0.01&0.144&-0.449&3.358&0.445&Yes&34379.422\\symb.\ $k=21$&Semi-Implicit&0.005&0.354&0.499&2.9&-0.499&Yes&10708.097\\symb.\ $k=21$&Semi-Implicit&0.005&-0.482&0.495&2.9&-0.499&Yes&10666.512\\symb.\ $k=21$&Semi-Implicit&0.005&0.067&-0.499&3.383&0.498&Yes&10829.929\\symb.\ $k=21$&Semi-Implicit&0.005&-0.226&-0.485&3.383&0.499&Yes&10742.843\\symb.\ $k=21$&Semi-Implicit&0.005&-0.282&-0.494&3.383&0.499&Yes&10788.866\\symb.\ $k=21$&Semi-Implicit&0.0025&0.097&-0.5&3.642&-0.5&Yes&11069.251\\symb.\ $k=21$&Semi-Implicit&0.0025&-0.291&-0.5&3.642&-0.5&Yes&11069.083\\symb.\ $k=21$&Semi-Implicit&0.0025&-0.329&0.5&2.642&0.5&Yes&11069.165\\symb.\ $k=21$&Semi-Implicit&0.0025&0.22&-0.5&3.642&-0.5&Yes&11069.087\\symb.\ $k=21$&Semi-Implicit&0.0025&-0.339&0.5&2.642&0.475&Yes&11045.46\\  
\hline
\end{tabular}
\caption{Proven persistent solutions for the cart-pole swingup model associated to the $k=21$ symbolic controller. }\label{cartpoleswing-persistent-21}
\end{table}

\end{document}